\documentclass[letterpaper]{article} 
\usepackage{aaai2026}  
\usepackage{times}  
\usepackage{helvet}  
\usepackage{courier}  
\usepackage[hyphens]{url}  
\usepackage{graphicx} 
\usepackage{natbib}  
\usepackage{caption} 
\usepackage{algorithm}
\usepackage{booktabs}
\usepackage{amsfonts}
\usepackage{nicefrac}
\usepackage{amsmath}
\usepackage{amssymb}
\usepackage{algpseudocode}
\usepackage{graphicx}
\usepackage{multirow}
\usepackage{array}
\usepackage{caption}
\usepackage{subcaption}
\usepackage{threeparttable}
\usepackage{colortbl}
\usepackage{xcolor}

\newtheorem{remark}{Remark}

\usepackage{newfloat}
\usepackage{listings}
\DeclareCaptionStyle{ruled}{labelfont=normalfont,labelsep=colon,strut=off} 
\floatstyle{ruled}
\newfloat{listing}{tb}{lst}{}
\floatname{listing}{Listing}
\title{Explore and Establish Synergistic Effects Between Weight Pruning and Coreset Selection in Neural Network Training}

\author {
    Weilin Wan\textsuperscript{\rm 1},
    Fan Yi\textsuperscript{\rm 1},
    Weizhong Zhang\textsuperscript{\rm 2}\thanks{Corresponding Author.},
    Quan Zhou\textsuperscript{\rm 1},
    Cheng Jin\textsuperscript{\rm 1,\rm 3}
}
\affiliations {
    \textsuperscript{\rm 1}College of Computer Science and Artificial Intelligence, Fudan University\\
    \textsuperscript{\rm 2}School of Data Science, Fudan University\\
    \textsuperscript{\rm 3}Shanghai Key Laboratory of Intelligent Information Processing\\
    \{wlwan23, fyi21\}@m.fudan.edu.cn, weizhongzhang@fudan.edu.cn,\\ qzhou21@m.fudan.edu.cn, jc@fudan.edu.cn
}

\usepackage{bibentry}

\begin{document}

\maketitle

\begin{abstract}
Modern deep neural networks rely heavily on massive model weights and training samples, incurring substantial computational costs.  
Weight pruning and coreset selection are two emerging paradigms proposed to improve computational efficiency.
In this paper, we first explore the interplay between redundant weights and training samples through a transparent  analysis: redundant samples, particularly noisy ones, cause model weights to become unnecessarily overtuned to fit them, complicating the identification of irrelevant weights during pruning; conversely, irrelevant weights tend to overfit noisy data, undermining coreset selection effectiveness.
To further investigate and harness this interplay in deep learning, we develop a \textbf{S}imultaneous \textbf{W}eight \textbf{a}nd \textbf{S}ample \textbf{T}ailoring mechanism (SWaST) that alternately performs weight pruning and coreset selection to establish a synergistic effect in training.
During this investigation, we observe that when simultaneously removing a large number of weights and samples, a phenomenon we term critical double-loss can occur, where important weights and their supportive samples are mistakenly eliminated at the same time, leading to model instability and nearly irreversible degradation that cannot be recovered in subsequent training. 
Unlike classic machine learning models, this issue can arise in deep learning due to the lack of theoretical guarantees on the correctness of weight pruning and coreset selection, which explains why these paradigms are often developed independently. 
We mitigate this by integrating a state preservation mechanism into SWaST, enabling stable joint optimization.  
Extensive experiments reveal a strong synergy between pruning and coreset selection across varying prune rates and coreset sizes, delivering accuracy boosts of up to 17.83\% alongside 10\% to 90\% FLOPs reductions.
\end{abstract}

\section{Introduction}
\label{sec:intro}
Training modern deep neural networks requires huge computational power and storage due to the large amount of model weights and datasets.
Weight pruning and coreset selection \citep{hoefler2021sparsity,killamsetty2021grad} are two emerging paradigms to enhance training efficiency or model quality. Their key ideas are to develop proper rules to remove redundant model weights and to select the most informative samples during training, respectively. 
Notably, in scenarios with noisy data, coreset selection has been shown to improve model accuracy by filtering out less valuable samples.
It has been repeatedly reported in the literature \citep{hoefler2021sparsity,mirzasoleiman2020coresets,killamsetty2021grad} that the proposed algorithms under these two paradigms can effectively reduce training costs with negligible performance degradation. 

However, we notice that weight pruning and coreset selection are always investigated independently and their interaction is overlooked. In contrast, for the two corresponding techniques i.e., feature screening \footnote{Features in classical machine learning models always have one-to-one correspondence  with model weights.} and sample screening, in classical machine learning such as support vector machines \cite{shibagaki2016simultaneous,zhang2017scaling}, their interaction has been extensively studied and exploited to develop efficient training acceleration algorithms. Theoretical analysis shows that there exists a significant synergistic effect between feature and sample screening, that is, discarding the identified redundant features (resp. samples)  leads to a more accurate estimation of the primal (resp. dual) optimum, which in turn enhances the sample (resp. feature) screening rules in detecting irrelevant samples (resp. features). These results are derived mainly depending on the theoretical tools, e.g., Karush–Kuhn–Tucker (KKT) conditions \cite{kuhn2013nonlinear}, in convex optimization. Effective joint screening methods are proposed to achieve a synergistic effect, which enables model training with 10 million of features and samples possible on the resource-restricted devices such as laptop.  {\it Although the above optimal tools do not exist in deep learning due to the highly complicated nonconvex training problem, there seems to be no reason to assert that a similar mechanism of the synergy effect does not hold in weight pruning and coreset selection paradigms, which is the main motivation of this work. }

The first thing that can be expected is that the coreset selection training process appears to be inherently more susceptible to overfitting and thus would benefit more from the regularization effects of weight pruning.
To provide an in-depth and transparent analysis of the interaction between weight pruning and coreset selection in deep learning, and considering the complexity of deep neural networks, we first perform an in-depth investigation on polynomial interpolation tasks. 
We employ standard weight pruning and coreset selection techniques for neural networks in this experiment (Fig. \ref{fig:poly}) to ensure that the results reflect deep learning principles. 
We get the following two key observations. One is that an excessive number of samples to be interpolated, especially noisy ones, can cause the coefficients of the polynomial to be overtuned in order to fit all the samples. This significantly complicates the identification of irrelevant weights during pruning, as most existing pruning methods develop their importance metric explicitly or implicitly based on the weight magnitude. The other observation is that, under the standard metric in coreset selection, the difficulty of selection grows exponentially with the number of redundant model weights. Our initial findings expose a critical interaction between model weights and training samples: redundant samples complicate weight pruning, while irrelevant weights obscure coreset selection. 
This mutual interference implies the limitations of traditional approaches that treat pruning and coreset selection as independent processes.

To further investigate and harness the above interplay in deep learning, we develop a \textbf{S}imultaneous \textbf{W}eight \textbf{a}nd \textbf{S}ample \textbf{T}ailoring mechanism (SWaST) that alternately performs weight pruning and coreset selection to establish a synergistic effect in training.
Every \(\mathcal{R}\) epochs, SWaST performs coreset selection to identify the most representative samples. During subsequent \(\mathcal{R}\) epochs, it performs training and online pruning using this selected coreset.
We derive two variants based on the aggressiveness of the pruning strategy: SWaST-trim (which prunes only the fully connected layers) and SWaST-cut (which prunes the entire network). SWaST-trim utilizes a conservative pruning strategy, retaining more parameters and ensuring better training stability. In contrast, SWaST-cut removes more parameters for greater efficiency, but this can lead to training instability, referred to as “{\it critical double-loss}”. This instability occurs when pivotal weights and their supportive samples are mistakenly eliminated at the same time, causing an almost irreversible performance degradation. 
In contrast to classic machine learning models, this phenomenon can easily occur in deep learning due to the absence of theoretical guarantees in pruning and coreset selection. This underlying complexity elucidates why existing methods in pruning and coreset selection have traditionally been developed in isolation.
To ensure training stability, we incorporate a state preservation mechanism that captures the model's output following coreset selection. This mechanism helps preserve critical knowledge during training and pruning while allowing adaptation to the refined dataset and architecture. Extensive experiments reveal a strong synergy between pruning and coreset selection across varying prune rates and coreset sizes, delivering accuracy boosts of up to 17.83\% alongside 10\% to 90\% FLOPs reductions.

Our main contributions can be summarized as follows.
\begin{itemize}
\item We conduct a transparent analysis to explore and identify the fundamental interplay between weight pruning and coreset selection within deep learning.
\item We propose an alternative pruning and coreset selection method SWaST, designed to further investigate and  exploit the interplay to establish a synergistic effect.
\item We design a state preservation mechanism for SWaST that stabilizes training and maintains model performance under aggressive pruning rates and small coreset sizes.
\item Experiments verify the synergistic effect established by SWaST, with a significant accuracy gain of up to 17.83\%.
\end{itemize}
\section{Related Work}  
\subsection{Coreset Selection}  
Coreset selection identifies a representative subset from the original (noisy) dataset with the objective of increasing training efficiency or enhancing model effectiveness. Recent methods are categorized into optimization-based and heuristic-based. Optimization-based approaches employ bilevel optimization~\cite{borsos2020coresets,zhou2022probabilistic,hao2024bilevel} or gradient matching objectives~\cite{killamsetty2021grad,mirzasoleiman2020coresets,killamsetty2021glister,killamsetty2021retrieve}, providing theoretical guarantees, but incurring high costs. Heuristic-based methods assess sample importance via forgetting events~\cite{toneva2018empirical}, gradient~\cite{pruthi2020estimating,xia2024less,thakkar2023self}, and inter-sample distances~\cite{xia2022moderate}. Recent progress includes DNN-based selectors~\cite{ilyas2022datamodels,engstrom2024dsdm} and efficient frameworks~\cite{everaert2023gio,paul2021deep,xiao2024feature,park2024robust} that balance quality and efficiency.

\begin{figure*}[!htbp]
  \centering
  \begin{subfigure}{0.32\linewidth}
    \includegraphics[width=1.0\linewidth]{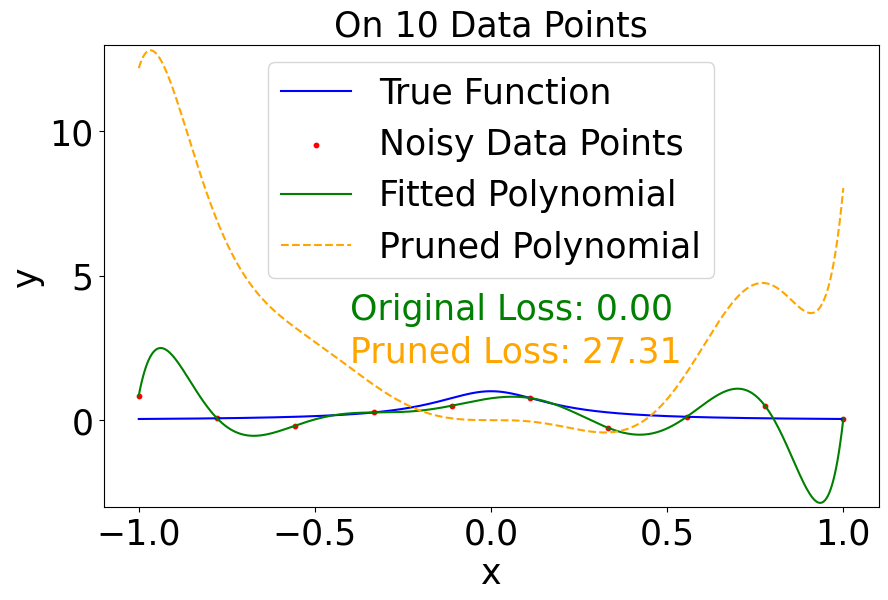}
    \caption{On noisy samples}
    \label{fig:fit_10}
  \end{subfigure}
  \hfill
  \begin{subfigure}{0.32\linewidth}
    \includegraphics[width=1.0\linewidth]{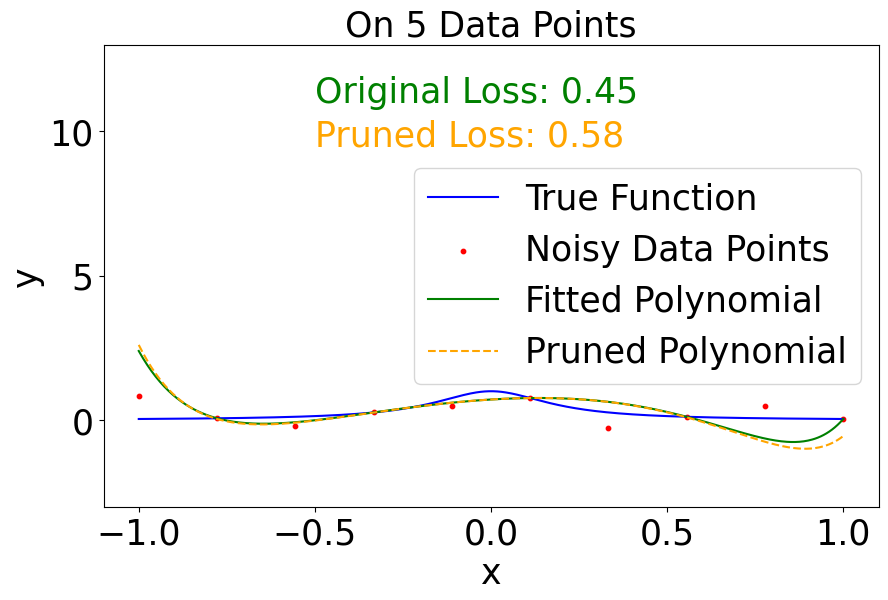}
    \caption{On clean samples}
    \label{fig:fit_5}
  \end{subfigure}
  \hfill
  \begin{subfigure}{0.32\linewidth}
    \includegraphics[width=1.0\linewidth]{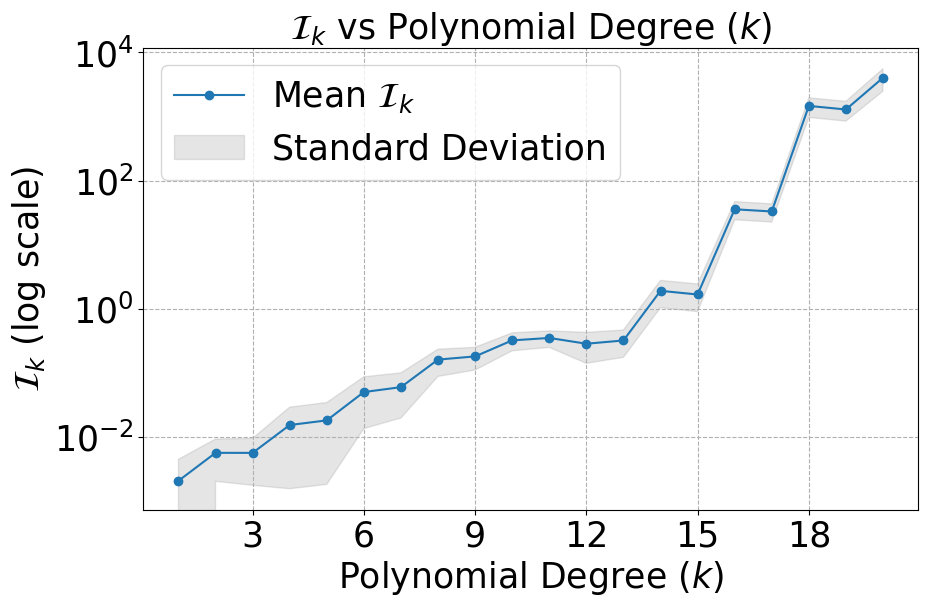}
    \caption{\( \mathcal{I}(\mathcal{D}, \hat{\mathcal{D}}) \) values}
    \label{fig:epsilon}
  \end{subfigure}
  \caption{\textbf{The impact of redundant weights and samples on pruning and coreset selection.} 
  The collapsed pruned model ({\color{orange} yellow curve}) in (a) compared to (b) implies that noisy/redundant data increases the difficulty of pruning.
  (c) shows selection difficulty \( \mathcal{I}(\mathcal{D}, \hat{\mathcal{D}}) \) rises with polynomial degree, indicating harder coreset selection.}
  \label{fig:poly}
\end{figure*}

\subsection{Weight Pruning}  
Weight pruning is a key technique for model compression and acceleration, classified into structured and unstructured methods. Structured pruning methods~\cite{filterspruning,he2018soft,jiang2022channel,elkerdawy2022fire,guan2022dais,nonnenmacher2021sosp,alvarez2016learning,murray2015auto} remove entire filters or channels, enabling hardware acceleration but often reducing accuracy. In contrast, unstructured pruning~\cite{tanaka2020pruning,lee2018snip,han2015deep,frankle2018lottery} eliminates individual weights based on metrics like magnitude or gradient sensitivity~\cite{evci2020rigging,han2015learning,wen2016learning,blakeney2020pruning,renda2020comparing,hassibi1992second,lecun1989optimal}. Recent theoretical work link pruning to optimization theory~\cite{grosse2016kronecker,zhou2021effective,srinivas2017training,louizos2017learning,qian2021probabilistic}, bolstering its foundations.

\subsection{Pruning and Coreset Selection in Classical ML}
In classical machine learning, feature and sample screening mirror weight pruning and coreset selection. Optimization-based theoretical frameworks exist for joint feature and sample screening in this domain~\cite{shibagaki2016simultaneous,zhang2017scaling}. Empirically, feature screening enhances sample importance estimation, and vice versa. While these theories don't translate directly to deep learning, this synergy motivates integrating weight pruning and coreset selection.
\section{Interplay of Redundant Samples and Weights}
\label{sec:poly}
We explore the interplay between redundant weights and samples by investigating their impact on weight pruning and coreset selection. 
In order to provide mathematically transparent analysis, we apply standard pruning and coreset selection techniques to polynomial interpolation.

\paragraph{Impact of redundant samples on weight pruning.} We study this impact on the polynomial interpolation task with a classic example  
Runge's function \( f(x) = \frac{1}{1 + 25x^2}, \, x \in [-1, 1] \)~\cite{runge1901empirische}, which is smooth and has derivatives of all orders, as shown by the {\color{blue}blue curve} in Fig.~\ref{fig:fit_10} and~\ref{fig:fit_5} (see appendix for results on other functions).
Our dataset comprises 10 points sampled from the function: 5 are clean data points (which form the coreset), and the other 5 are noisy points, generated by adding Gaussian noise with a magnitude of 0.1. Using the least squares method, we fitted a \(10^{\text{th}}\)-degree polynomial to two datasets separately: the full dataset of 10 points and the 5-point coreset. Following the standard weight pruning techniques~\cite{han2016deep}, we pruned the three smallest coefficients (in magnitude) by setting them to zero while retaining the remaining coefficients. 
Our findings indicate that an increased number of data points, especially those containing noise, renders pruning more challenging. This is evidenced by the higher loss observed ({\color{orange}yellow curve} in Fig.~\ref{fig:fit_10}) after pruning, which suggests that coreset selection can enhance the effectiveness of pruning in scenarios involving redundant samples.

\paragraph{Impact of redundant weights on coreset selection.}
The goal of coreset selection is to identify a small subset of training data that approximates the objective function of the entire dataset throughout the parameter space. A typical formulation is constructed on function approximation, which involves finding a small subset \( \hat{\mathcal{D}} \) from the training dataset \( \mathcal{D} \) such that:
\begin{align}
   \frac{|\mathcal{L}(\boldsymbol{\theta}) - \hat{\mathcal{L}}(\boldsymbol{\theta})|}{\mathcal{L}(\boldsymbol{\theta})} \leq \epsilon , ~\text{for any } \boldsymbol{\theta} \in \mathbb{R}^p, \label{eq:epsilon}
\end{align}
where \( \mathcal{L}(\boldsymbol{\theta}) = \frac{1}{|\mathcal{D}|} \sum_{i \in \mathcal{D}} \ell(\boldsymbol{\theta}; x_i, y_i) \) is the objective function on the full dataset, and \( \hat{\mathcal{L}}(\boldsymbol{\theta}) = \frac{1}{|\hat{\mathcal{D}}|} \sum_{i \in \hat{\mathcal{D}}} \ell(\boldsymbol{\theta}; x_i, y_i) \) is the objective function on the selected subset \( \hat{\mathcal{D}} \). Here, \( \epsilon \) is a small number that controls the approximation error. When \( \epsilon \) is sufficiently small, the model performance learned from \( \hat{\mathcal{L}} \) can be comparable to that learned from \( \mathcal{L} \). In this context, 
\begin{align}
   \mathcal{I}(\mathcal{D}, \hat{\mathcal{D}}) = \sup_{\boldsymbol{\theta}}\frac{|\mathcal{L}(\boldsymbol{\theta}) - \hat{\mathcal{L}}(\boldsymbol{\theta})|}{\mathcal{L}(\boldsymbol{\theta})}, \label{eq:idd}
\end{align}
can be used to evaluate the difficulty of coreset selection. The reason is that given two fixed datasets $\mathcal{D}$ and $\hat{\mathcal{D}}$, if $\mathcal{I}(\mathcal{D}, \hat{\mathcal{D}}) $ for network $f_1(\cdot, \boldsymbol{\theta})$ is larger than network $f_2(\cdot, \boldsymbol{\theta})$, it means that $f_1(\cdot, \boldsymbol{\theta})$ needs to choose a larger coreset $\hat{\mathcal{D}}$ to achieve the same accuracy with $f_1(\cdot, \boldsymbol{\theta})$, i.e., the coreset selection in $f_1(\cdot, \boldsymbol{\theta})$ is more difficult than $f_2(\cdot, \boldsymbol{\theta})$.

In our second polynomial experiment, we employed Runge's function to create samples and applied Gaussian noise with a strength of 0.1. We optimized the problem in Eq.~(\ref{eq:idd}) for polynomial degrees \( k = 1, \ldots, 20 \) and present the result in Fig.~\ref{fig:epsilon}. It shows that with the y-axis in logarithmic scale, \( \mathcal{I}(\mathcal{D}, \hat{\mathcal{D}}) \) grows exponentially with increasing polynomial degree. The reason is that $\mathcal{L}(\boldsymbol{\theta})$ vanishes to zero rapidly due to the overfitting of the noisy data. 
This suggests that coreset selection difficulty increases with model parameter dimension rapidly. Therefore, it can be expected that pruning can mitigate this issue by reducing model size.

\begin{figure*}[!htbp]
  \centering
  \includegraphics[width=0.75\linewidth]{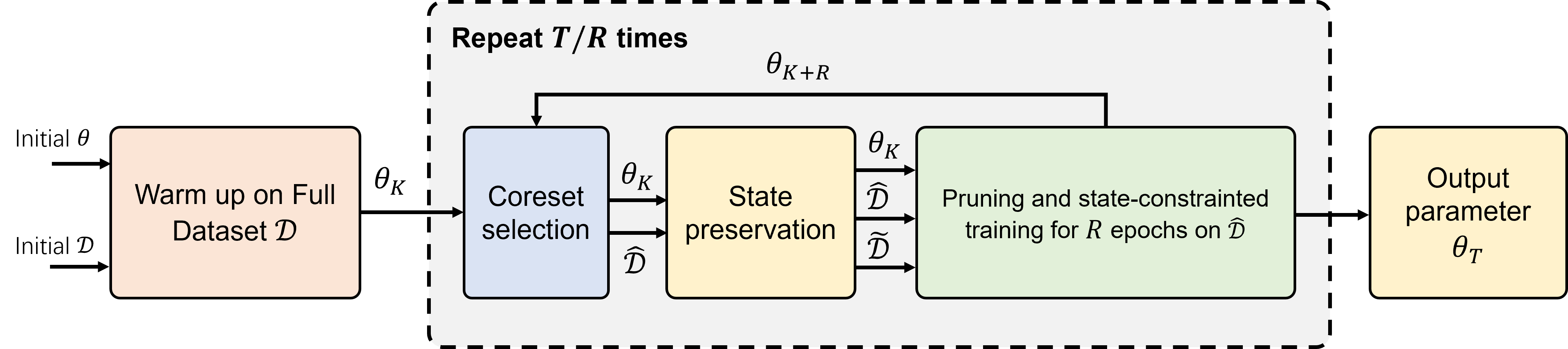}
   \caption{Overview of SWaST, alternating between pruning and coreset selection every \(\mathcal{R}\) epochs. \(\mathcal{D}\) denotes the full dataset, \(\hat{\mathcal{D}}\) denotes the selected subset, \(\tilde{\mathcal{D}}\) denotes the stored logits, and \(T\) denotes the total epochs.}
   \label{fig:mechanism}
\end{figure*}

\section{The Proposed Training Mechanism} 
In this section, we propose SWaST to further investigate and exploit the interplay above to establish a synergistic effect between weights pruning and coreset selection.

\subsection{A Synergistic Optimization Mechanism}
\label{sec:3-1}
SWaST consists of a warm-up phase and an alternating optimization phase (shown in Fig.~\ref{fig:mechanism}). 
In the warm-up phase, the network is trained on the full dataset for \(\mathcal{K}\) epochs to establish a good initialization. 
The subsequent phase alternatively performs weight pruning and coreset selection, where coreset selection is performed every \(\mathcal{R}\) epochs to identify representative samples,
followed by network training on the selected samples with pruning operations. This alternating process effectively removes redundant samples and parameters while maintaining model performance. The detailed procedure is outlined in Algorithm~\ref{alg:mechanism}.
We propose two variants that use different pruning strategies: 
\begin{itemize}
    \item[(i)] \textbf{SWaST-trim} prunes only the FC layer, retaining most parameters to ensure training stability. The improved efficiency is mainly attributed to coreset selection.
    \item[(ii)] \textbf{SWaST-cut} prunes the entire network to achieve more aggressive efficiency gains, but may lead to training instability (which we refer to as the \textbf{“double-loss problem"}). We develop the state preservation mechanism to mitigate this issue in the following section.
\end{itemize}

\begin{algorithm}[!h]  
\caption{SWaST}  
\label{alg:mechanism}  
\begin{algorithmic}[1]  
\Require  Dataset $\mathcal{D}$, warm-up epochs $\mathcal{K}$, epoch interval $\mathcal{R}$ for coreset selection, coreset to full set ratio $\alpha$, pruning algorithm $\mathcal{P}$, coreset selection algorithm $\mathcal{S}$, state preservation weight $\lambda$, use state preservation flag $use\_SP$.  
\Ensure Trained network parameters $\boldsymbol{\theta}$  

\State Initialize network parameters $\boldsymbol{\theta}$  
\State Set $\mathcal{X} = \mathcal{D}$  
\State Initialize stored logits $\tilde{\mathcal{D}} = \{\}$  
\For{$t = 1, 2, \ldots, T$}  
    \If{$t > \mathcal{K}$ and $t \mod \mathcal{R} = 0$}  
        \State $\mathcal{X} \gets \mathcal{S}(\mathcal{D}, \boldsymbol{\theta}, \alpha)$ \Comment{Execute coreset selection}
        \If{$use\_SP$}
            \State $\tilde{\mathcal{D}} \gets \{(\mathbf{x}_i, \mathbf{z}_i):
\mathbf{z}_i = f_{\boldsymbol{\theta}_{\textit{pre}}}(\mathbf{x}_i),  \mathbf{x}_i \in \mathcal{X} \}$
        \EndIf
    \EndIf  
    \For{each training iteration}  
        \State Sample mini-batch $\mathcal{B}$ from coreset $\mathcal{X}$  
        \State Compute cross-entropy loss $\mathcal{L}_{CE}$  
        \If{$\mathbf{Z}$ is not empty}  
            \State Compute $\mathcal{L}_{SP}$ using $\tilde{\mathcal{D}}$   
            \State $\mathcal{L}_{\text{total}} = \mathcal{L}_{CE} + \lambda\mathcal{L}_{SP}$ \Comment{SWaST-cut} 
        \Else  
            \State $\mathcal{L}_{\text{total}} = \mathcal{L}_{CE}$ \Comment{SWaST-trim} 
        \EndIf  
        \State Compute gradients $\nabla_{\boldsymbol{\theta}} \mathcal{L}_{\text{total}}$   
        \State $\boldsymbol{\theta} \gets \mathcal{P}(\boldsymbol{\theta}, \nabla_{\boldsymbol{\theta}} \mathcal{L}_{\text{total}})$ \Comment{Execute pruning step}   
        \State Update parameters $\boldsymbol{\theta}$ using gradient descent  
    \EndFor  
\EndFor  
\State \Return $\boldsymbol{\theta}$  
\end{algorithmic}  
\end{algorithm}

\begin{remark}
    SWaST maintains generality by supporting any pruning algorithm and coreset selection method that can be integrated into the training process. 
    \begin{itemize}
        \item For pruning, the only requirement is that the method should be capable of operating in an online manner, updating the network structure during training rather than requiring separate pre-training or post-processing steps.
        \item For coreset selection, any method that evaluates and selects samples based on the current model is compatible.
    \end{itemize} 
    This flexibility allows practitioners to select the most suitable pruning and coreset selection approaches for their specific needs while still benefiting from SWaST's synergistic optimization between pruning and coreset selection.
\end{remark}

\subsection{Double-loss Problem: Analysis and Mitigation}
We begin by establishing two baseline observations:
\begin{itemize}
    \item For pruning only scenarios, when a parameter is incorrectly pruned, it can be recovered through learning from appropriate training samples in the next iteration.
    \item Similarly, for coreset only scenarios, when a training sample is mistakenly excluded, the knowledge preserved in the network parameters can help reidentify and select this sample in the next round.
\end{itemize}

\begin{figure}[!t]
  \centering
  \includegraphics[width=1.0\linewidth]{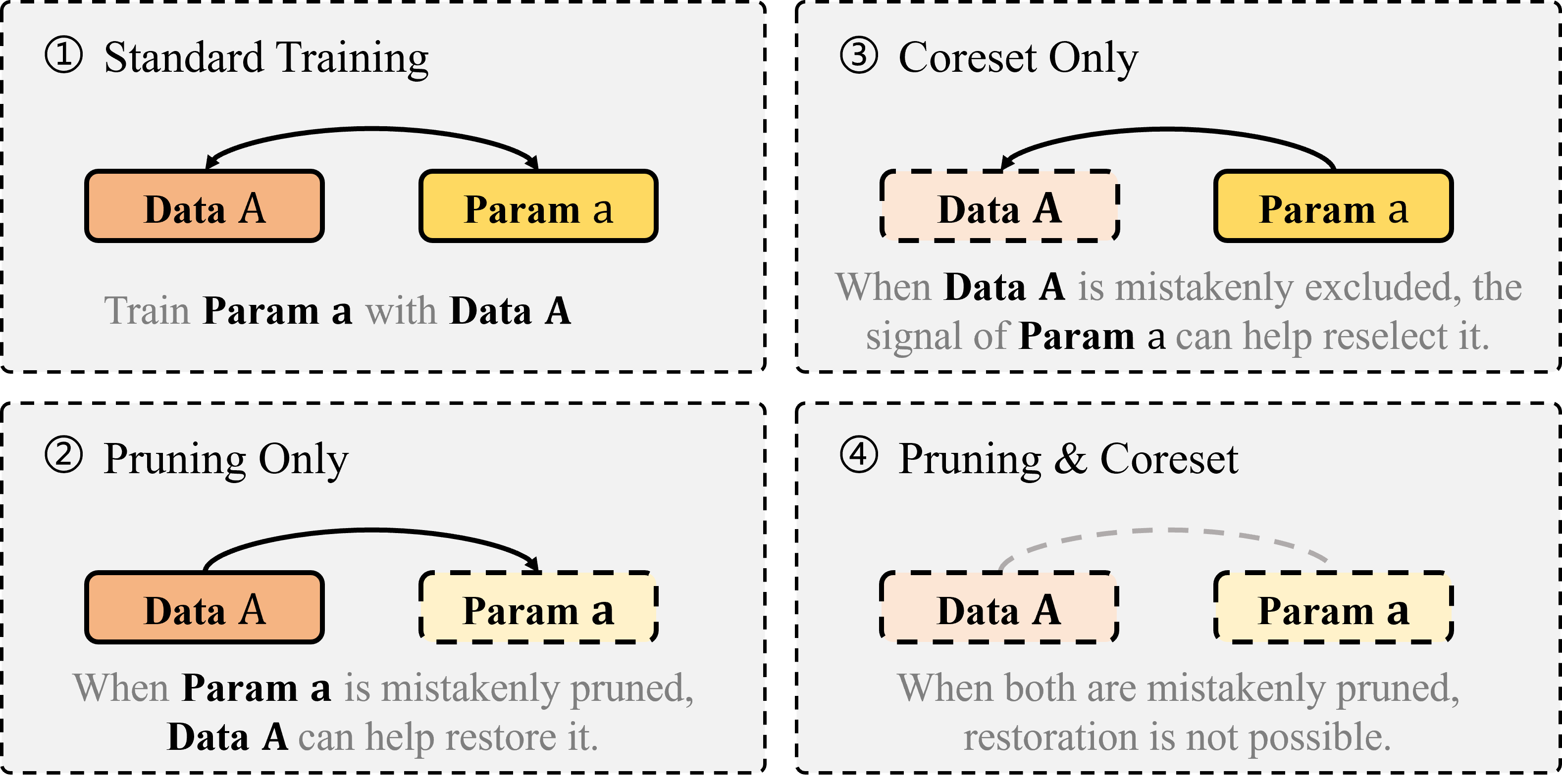}
   \caption{Illustration of the “critical double-loss" phenomenon in concurrent weight pruning and sample tailoring: (1) Standard Training shows the close link between Data $A$ and Param $a$. (2) Pruning Only allows recovery of Param $a$ using Data $A$ when mistakenly pruned. (3) Coreset Only supports re-selection of Data $A$ through Param $a$ when excluded in error. (4) Pruning \& Coreset causes irreversible degradation from simultaneous exclusion.}
   \label{fig:double-loss}
\end{figure}

However, when performing weight pruning and sample tailoring simultaneously, we may encounter situations in which both a parameter and its corresponding supportive training samples are eliminated together. In such cases, standard training procedures struggle to recover from this information loss, as neither component remains available to assist in the recovery process. We term this the “double-loss" problem. Fig.~\ref{fig:double-loss} illustrates this phenomenon through a representative pair of interrelated data and parameters.

\begin{remark}
    Fig.~\ref{fig:double-loss} offers an intuitive illustration of the double-loss phenomenon, rather than experimental or theoretical validation. It depicts how losing interconnected parameters and samples can impede training recovery. The actual parameter-sample interactions in neural networks occur in higher dimensions with greater complexity.
\end{remark}

\begin{remark}  
    It is important to note that the critical double-loss phenomenon is unique to deep learning. Classical machine learning models leverage theoretical tools like KKT conditions ~\cite{shibagaki2016simultaneous} for safe feature/sample removal. In contrast, deep models' highly non-convex loss function prevent such guarantees, necessitating dynamic stabilization mechanisms, which is developed below. 
\end{remark}

To address this challenge while maintaining optimization stability, we propose \textbf{a state preservation mechanism} that operates through two alternating phases:

\textbf{Stage 1: State recording} (executed every $\mathcal{R}$ epochs). Following each coreset update, we capture the model's states via full forward propagation, i.e.,  
$$\tilde{\mathcal{D}}=\{(\mathbf{x}_i, \mathbf{z}_i):
\mathbf{z}_i = f_{\boldsymbol{\theta}_{\textit{pre}}}(\mathbf{x}_i),  \mathbf{x}_i \in \mathcal{X}  \}.$$  
These preserved states encode critical information about the current model and data subset, including both explicit prediction patterns and feature correlations.  

\textbf{Stage 2: State-constrained training}. During subsequent parameter updates, we enforce state consistency through the composite loss:  
\begin{equation}
    \label{eq:loss}
    \mathcal{L}_{\text{total}} = \mathcal{L}_{\text{CE}} + \lambda\!\!\!\!\!\sum_{(\mathbf{x}_i,\mathbf{z}_i ) \in \tilde{\mathcal{D}}}\!\!\!\!\!\text{KL}(\sigma(\mathbf{z}_i) \| \sigma(f_{\boldsymbol{\theta}}(\mathbf{x}_i))), 
\end{equation}   
where $\lambda$ (default 0.1) balances the primary learning objective with state consistency, $\sigma$ denotes the softmax function, and KL represents the Kullback-Leibler divergence.
Note that the KL terms in Eq.(\ref{eq:loss}) imply that we utilize all samples in \(\mathcal{X}\), as even misclassified examples provide valuable information about the model’s optimization trajectory.  

The mechanism above directly addresses the “double-loss” problem with the following capabilities:
\begin{itemize}
    \item KL terms can correct mistakes from previous pruning steps. If a critical parameter is erroneously pruned, the KL term can detect this error in subsequent steps—since the KL divergence relative to the preserved state \(\mathbf{z}_i\) will increase—and re-select that parameter. 
    \item The KL terms help mitigate training instability caused by pruning. Mistakenly removing important parameters can lead to large fluctuations in \(\sigma(\mathbf{z}_i)\) and destabilize training. Such fluctuations are detected by the KL term, which measures distributional divergence across all classes, whereas \(\mathcal{L}_{\mathrm{CE}}\) only penalizes the ground-truth class.
\end{itemize}

\section{Experiments}
\label{sec:exp}
To evaluate the effectiveness and robustness of SWaST, we conducted a series of experiments. All results reported in the tables are averaged over 5 runs. 
We first briefly review the experimental setup, followed by specific results. Detailed method descriptions and additional results with other pruning/coreset methods, can be found in the Appendix.

\begin{itemize}
    \item \textbf{Datasets and network architectures:} We conduct experiments on CIFAR-10/100 \cite{krizhevsky2009cifar}, TinyImageNet~\cite{le2015tiny}, and ImageNet-1K~\cite{deng2009imagenet} datasets, using ResNet-18 and ResNet-101 to verify our method's effectiveness across different model scales.  
    \item \textbf{Coreset selection and pruning methods:} Given the varying scales of datasets, we adopt selection methods with different computational complexities: GradMatch~\cite{killamsetty2021grad} for CIFAR, Moderate~\cite{xia2022moderate} for TinyImageNet, and EL2N~\cite{paul2021deep} for ImageNet-1K. We employ the RigL~\cite{evci2020rigging} method as our default pruning algorithm.
    \item \textbf{Pruning configuration:} For SWaST-trim, the pruning rate applies to FC layers; for SWaST-cut, it applies to the backbone with FC layers fixed at 90\% pruning. Configuration details are provided in the Appendix. Detailed analysis and justification for these configuration choices are provided in the Appendix.
    \item \textbf{Evaluation on datasets with label noise: } To thoroughly analyze the synergistic effects between weight pruning and coreset selection, we introduce label noise into the datasets in certain experiments.
\end{itemize}

\subsection{Main Results}
\label{sec:4-1}

\paragraph{Efficiency analysis.} 
\begin{figure}[!b]
  \centering
  \includegraphics[width=1.0\linewidth]{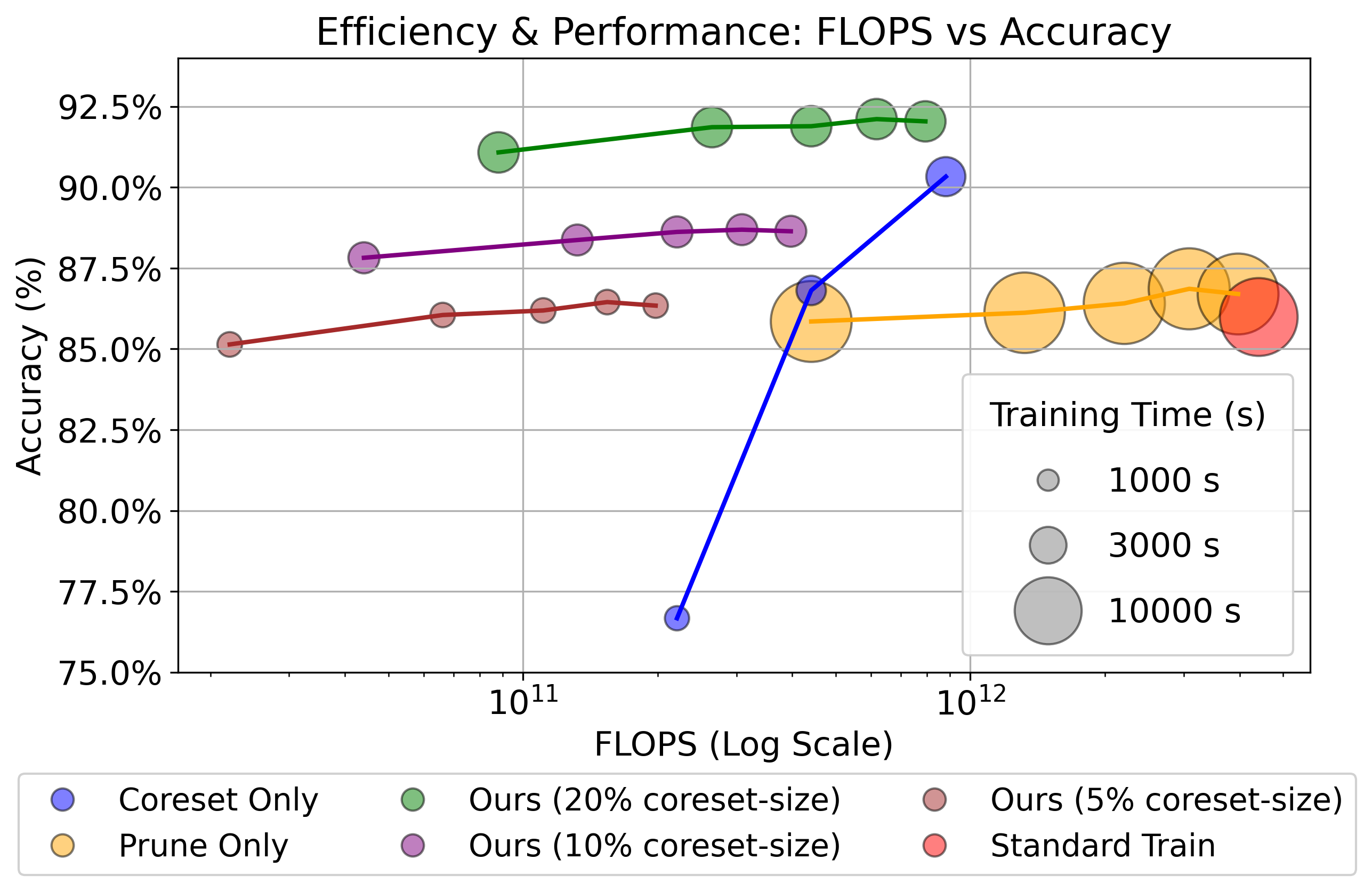}
   \caption{Efficiency vs.\ accuracy trade-offs (bubble area \(\propto\) training time), where the upper-left region indicates optimal performance (high accuracy with low FLOPs).}
   \label{fig:efficiency}
\end{figure}
\label{sec:efficiency_analysis}  

\begin{table*}[!t] 
    \centering  
    \resizebox{\textwidth}{!}{  
    \renewcommand{\arraystretch}{1.2}
    \begin{tabular}{lcccccccc}  
    \hline
    &  &  & \multicolumn{5}{c}{\textbf{SWaST-trim with different Prune Rate}} &  \\ \cline{4-8}  
    \multirow{-2}{*}{\textbf{Model}} & \multirow{-2}{*}{\textbf{Dataset}} & \multirow{-2}{*}{\textbf{Coreset Size}} & 90\% & 70\% & 50\% & 30\% & 10\% & \multirow{-2}{*}{\textbf{Coreset-only}} \\ \hline  
    &  & 10\% & {\color{red}$92.58_{{\tiny(+0.38)}}$} & {\color{blue}$92.53_{{\tiny(+0.33)}}$} & {\color{blue}$92.40_{{\tiny(+0.20)}}$} & $92.32_{{\tiny(+0.12)}}$ & $92.29_{{\tiny(+0.09)}}$ & 92.20 \\
    &  & 5\% & {\color{red}$90.16_{{\tiny(+0.49)}}$} & {\color{blue}$90.07_{{\tiny(+0.40)}}$} & {\color{blue}\textbf{$89.92_{{\tiny(+0.25)}}$}} & $89.81_{{\tiny(+0.14)}}$ & $89.77_{{\tiny(+0.10)}}$ & 89.67 \\
    & \multirow{-3}{*}{CIFAR-10} & 1\% & {\color{red}$78.60_{{\tiny(+2.97)}}$} & {\color{blue}$78.33_{{\tiny(+2.80)}}$} & {\color{blue}$77.14_{{\tiny(+1.61)}}$} & \textbf{$76.26_{{\tiny(+0.73)}}$} & $76.01_{{\tiny(+0.48)}}$ & 75.53 \\ \cline{2-9}   
    &  & 10\% & {\color{red}$71.94_{{\tiny(+0.97)}}$} & {\color{blue}$71.76_{{\tiny(+0.79)}}$} & {\color{blue}$71.44_{{\tiny(+0.47)}}$} & \textbf{$71.27_{{\tiny(+0.30)}}$} & $71.20_{{\tiny(+0.23)}}$ & 70.97 \\
    &  & 5\% & {\color{red}$66.19_{{\tiny(+1.40)}}$} & {\color{blue}$65.97_{{\tiny(+1.18)}}$} & {\color{blue}$65.56_{{\tiny(+0.77)}}$} & $65.22_{{\tiny(+0.43)}}$ & \textbf{$65.09_{{\tiny(+0.30)}}$} & 64.79 \\
    \multirow{-6}{*}{ResNet-18} & \multirow{-3}{*}{CIFAR-100} & 1\% & {\color{red}$38.38_{{\tiny(+1.96')}}$} & {\color{blue}$38.12_{{\tiny(+1.70)}}$} & {\color{blue}\textbf{$37.58_{{\tiny(+1.16)}}$}} & $37.01_{{\tiny(+0.59)}}$ & $36.88_{{\tiny(+0.46)}}$ & 36.42 \\ \hline  
    \multicolumn{1}{c}{} &  & 10\% & {\color{red}$92.60_{{\tiny(+1.47)}}$} & {\color{blue}$92.45_{{\tiny(+1.32)}}$} & {\color{blue}$92.12_{{\tiny(+0.99)}}$} & $91.89_{{\tiny(+0.76)}}$ & $91.54_{{\tiny(+0.41)}}$ & 91.13 \\
    \multicolumn{1}{c}{} &  & 5\% & {\color{red}$89.81_{{\tiny(+3.53)}}$} & {\color{blue}$89.39_{{\tiny(+3.11)}}$} & {\color{blue}$88.04_{{\tiny(+1.76)}}$} & $87.46_{{\tiny(+1.18)}}$ & $87.29_{{\tiny(+1.01)}}$ & 86.28 \\
    \multicolumn{1}{c}{} & \multirow{-3}{*}{CIFAR-10} & 1\% & {\color{red}$66.23_{{\tiny(+16.51)}}$} & {\color{blue}$63.36_{{\tiny(+13.64)}}$} & {\color{blue}$60.46_{{\tiny(+10.74)}}$} & $55.45_{{\tiny(+5.73)}}$ & $53.02_{{\tiny(+3.30)}}$ & 49.72 \\ \cline{2-9}   
    \multicolumn{1}{c}{} &  & 10\% & {\color{red}$71.93_{{\tiny(+3.14)}}$} & {\color{blue}$71.68_{{\tiny(+2.89)}}$} & {\color{blue}$70.47_{{\tiny(+1.68)}}$} & $70.01_{{\tiny(+1.22)}}$ & $68.74_{{\tiny(+0.95)}}$ & 68.79 \\
    \multicolumn{1}{c}{} &  & 5\% & {\color{red}$61.72_{{\tiny(+7.84)}}$} & {\color{blue}$60.25_{{\tiny(+6.37)}}$} & {\color{blue}$58.19_{{\tiny(+4.31)}}$} & $56.63_{{\tiny(+2.75)}}$ & $55.82_{{\tiny(+1.94)}}$ & 53.88 \\
    \multirow{-6}{*}{ResNet-101} & \multirow{-3}{*}{CIFAR-100} & 1\% & {\color{red}$21.16_{{\tiny(+9.47)}}$} & {\color{blue}$18.53_{{\tiny(+6.84)}}$} & {\color{blue}\textbf{$16.04_{{\tiny(+4.35)}}$}} & $14.46_{{\tiny(+2.77)}}$ & $13.87_{{\tiny(+2.18)}}$ & 11.69 \\ \hline
    &  &  & \multicolumn{5}{c}{\textbf{SWaST-cut with different Prune Rate}} &  \\ \cline{4-8}  
    \multirow{-2}{*}{\textbf{Model}} & \multirow{-2}{*}{\textbf{Dataset}} & \multirow{-2}{*}{\textbf{Coreset Size}} & 90\% & 70\% & 50\% & 30\% & 10\% & \multirow{-2}{*}{\textbf{Coreset-only}} \\ \hline  
    &  & 10\% & $91.53_{{\tiny(-0.67)}}$ & $92.32_{{\tiny(+0.12)}}$ & {\color{blue}$92.40_{{\tiny(+0.20)}}$} & {\color{red}\textbf{$92.92_{{\tiny(+0.72)}}$}} & {\color{blue}$92.60_{{\tiny(+0.40)}}$} & 92.20 \\
    &  & 5\% & $89.32_{{\tiny(-0.35)}}$ & $90.12_{{\tiny(+0.45)}}$ & {\color{blue}\textbf{$90.35_{{\tiny(+0.68)}}$}} & {\color{red}$91.11_{{\tiny(+1.44)}}$} & {\color{blue}$90.18_{{\tiny(+0.51)}}$} & 89.67 \\
    & \multirow{-3}{*}{CIFAR-10} & 1\% & $77.13_{{\tiny(+1.60)}}$ & $78.52_{{\tiny(+2.99)}}$ & {\color{blue}$78.57_{{\tiny(+3.24)}}$} & {\color{red}\textbf{$78.70_{{\tiny(+3.17)}}$}} & {\color{blue}$78.67_{{\tiny(+3.14)}}$} & 75.53 \\ \cline{2-9}   
    &  & 10\% & $67.98_{{\tiny(-2.99)}}$ & $71.09_{{\tiny(+0.12)}}$ & {\color{blue}$71.91_{{\tiny(+0.94)}}$} & {\color{red}\textbf{$72.09_{{\tiny(+1.12)}}$}} & {\color{blue}$71.99_{{\tiny(+1.02)}}$} & 70.97 \\
    &  & 5\% & $63.24_{{\tiny(-1.55)}}$ & $65.15_{{\tiny(+0.36)}}$ & {\color{blue}$65.55_{{\tiny(+0.76)}}$} & {\color{red}$66.81_{{\tiny(+2.02)}}$} & {\color{blue}\textbf{$66.30_{{\tiny(+1.51)}}$}} & 64.79 \\
    \multirow{-6}{*}{ResNet-18} & \multirow{-3}{*}{CIFAR-100} & 1\% & $35.80_{{\tiny(-0.62)}}$ & $37.81_{{\tiny(+1.39)}}$ & {\color{red}\textbf{$40.29_{{\tiny(+3.87)}}$}} & {\color{blue}$39.55_{{\tiny(+3.13)}}$} & {\color{blue}$38.53_{{\tiny(+2.11)}}$} & 36.42 \\ \hline
    \multicolumn{3}{c}{\textbf{Params. After Pruning}} & 1.2M & 3.5M & 5.8M & 8.2M & 10.5M & 11.7M \\ \hline
    
    \multicolumn{1}{c}{} &  & 10\% & $91.85_{{\tiny(+0.72)}}$ & $92.46_{{\tiny(+1.33)}}$ & {\color{blue}$92.61_{{\tiny(+1.48)}}$} & {\color{red}\textbf{$92.96_{{\tiny(+1.83)}}$}} & {\color{blue}$92.62_{{\tiny(+1.49)}}$} & 91.13 \\
    \multicolumn{1}{c}{} &  & 5\% & $88.90_{{\tiny(+2.62)}}$ & $89.55_{{\tiny(+3.27)}}$ & {\color{blue}$90.00_{{\tiny(+3.72)}}$} & {\color{red}\textbf{$90.13_{{\tiny(+3.85)}}$}} & {\color{blue}$89.99_{{\tiny(+3.71)}}$} & 86.28 \\
    \multicolumn{1}{c}{} & \multirow{-3}{*}{CIFAR-10} & 1\% & $64.57_{{\tiny(+14.85)}}$ & $64.82_{{\tiny(+15.10)}}$ & {\color{blue}$66.92_{{\tiny(+17.20)}}$} & {\color{red}\textbf{$67.55_{{\tiny(+17.83)}}$}} & {\color{blue}$66.52_{{\tiny(+16.80)}}$}  & 49.72 \\ \cline{2-9}   
    \multicolumn{1}{c}{} &  & 10\% & $69.41_{{\tiny(+0.62)}}$ & $70.73_{{\tiny(+1.94)}}$ & {\color{blue}$71.75_{{\tiny(+2.96)}}$} & {\color{blue}$71.95_{{\tiny(+3.16)}}$} & {\color{red}\textbf{$72.05_{{\tiny(+3.26)}}$}} & 68.79 \\
    \multicolumn{1}{c}{} &  & 5\% & $59.27_{{\tiny(+5.39)}}$ & $59.65_{{\tiny(+5.77)}}$ & {\color{blue}$62.87_{{\tiny(+8.99)}}$} & {\color{red}\textbf{$62.97_{{\tiny(+9.09)}}$}} & {\color{blue}$62.14_{{\tiny(+8.26)}}$} & 53.88 \\
    & \multirow{-3}{*}{CIFAR-100} & 1\% & $19.17_{{\tiny(+7.48)}}$ & $20.11_{{\tiny(+8.42)}}$ & {\color{red}\textbf{$21.24_{{\tiny(+9.55)}}$}} & {\color{blue}$20.99_{{\tiny(+9.30)}}$} & {\color{blue}$21.19_{{\tiny(+9.50)}}$} & 11.69 \\ \cline{2-9}
    &  & 10\% & $51.29_{{\tiny(+2.63)}}$ & $52.78_{{\tiny(+4.12)}}$ & {\color{red}$53.49_{{\tiny(+4.83)}}$} & {\color{blue}$53.14_{{\tiny(+4.48)}}$} & {\color{blue}$52.98_{{\tiny(+4.32)}}$} & 48.66 \\
     & \multirow{-2}{*}{TinyImageNet} & 5\% & $38.65_{{\tiny(+3.55)}}$ & {\color{blue}$42.83_{{\tiny(+7.73)}}$} & {\color{red}$42.93_{{\tiny(+7.83)}}$} & {\color{blue}$41.94_{{\tiny(+6.84)}}$} & $41.21_{{\tiny(+6.11)}}$ & 35.10 \\ \cline{2-9}
    &  & 10\% & $37.55_{{\tiny(+5.83)}}$ & $38.28_{{\tiny(+6.56)}}$ & {\color{blue}$38.92_{{\tiny(+7.20)}}$} & {\color{blue}$39.19_{{\tiny(+7.47)}}$} & {\color{red}$39.47_{{\tiny(+7.75)}}$} & 31.72 \\
    \multirow{-10}{*}{ResNet-101} & \multirow{-2}{*}{ImageNet-1k} & 5\% & $31.63_{{\tiny(+1.35)}}$ & $32.25_{{\tiny(+1.97)}}$ & {\color{blue}$32.71_{{\tiny(+2.43)}}$} & {\color{red}$34.34_{{\tiny(+4.06)}}$} & {\color{blue}$33.12_{{\tiny(+2.84)}}$} & 30.28 \\ \hline
    \multicolumn{3}{c}{\textbf{Params. After Pruning}} & 4.5M & 13.3M & 22.3M & 31.2M & 40.1M & 44.5M \\ \hline
    \end{tabular}} 
    \caption{Comparison for various pruning rates and subset fractions on different datasets. Values in parentheses show differences from the Coreset-only baseline. {\color{red}Red}/{\color{blue}blue} numbers indicate best and runner-up results within each row.  SWaST-trim (upper half) prunes only FC layers, maintaining parameters near originals: $\sim$11.2M--11.7M for ResNet-18 (original 11.7M) and $\sim$42.7M--44.5M for ResNet-101 (original 44.5M). SWaST-cut (lower half) applies global pruning with explicit parameter counts listed.  \textbf{Row-wise}: Both SWaST variants consistently outperform the baseline, with gains up to \textbf{17.83\%}. 
    \textbf{Column-wise}: Performance improvements \textbf{increase as coreset sizes decrease} since smaller coresets present greater selection challenge and higher overfitting risk, while weight
    pruning mitigates these issues.
    }
    \label{tab:tab1}
\end{table*}

To validate the efficiency of SWaST, we present experimental results conducted on ResNet101, as illustrated in Fig.~\ref{fig:efficiency}. In this visualization, the x-axis represents FLOPS, the y-axis shows accuracy, and the circle sizes indicate training time (detailed experimental configurations and settings are provided in the appendix). As observed from the results, our method achieves significantly higher accuracy compared to other approaches under equivalent FLOPS constraints, demonstrating superior computational efficiency.

\paragraph{Weight pruning improves coreset selection.}
Following the protocol established in~\cite{killamsetty2021grad},  we benchmark our proposed methods (SWaST-trim and SWaST-cut) against the coreset-only baseline. Table~\ref{tab:tab1} demonstrates that SWaST variants exhibit two key advantages when operating with identical subset sizes:
\begin{figure*}[!htbp]
  \centering
  \begin{subfigure}{0.32\linewidth}
    \includegraphics[width=1.0\linewidth]{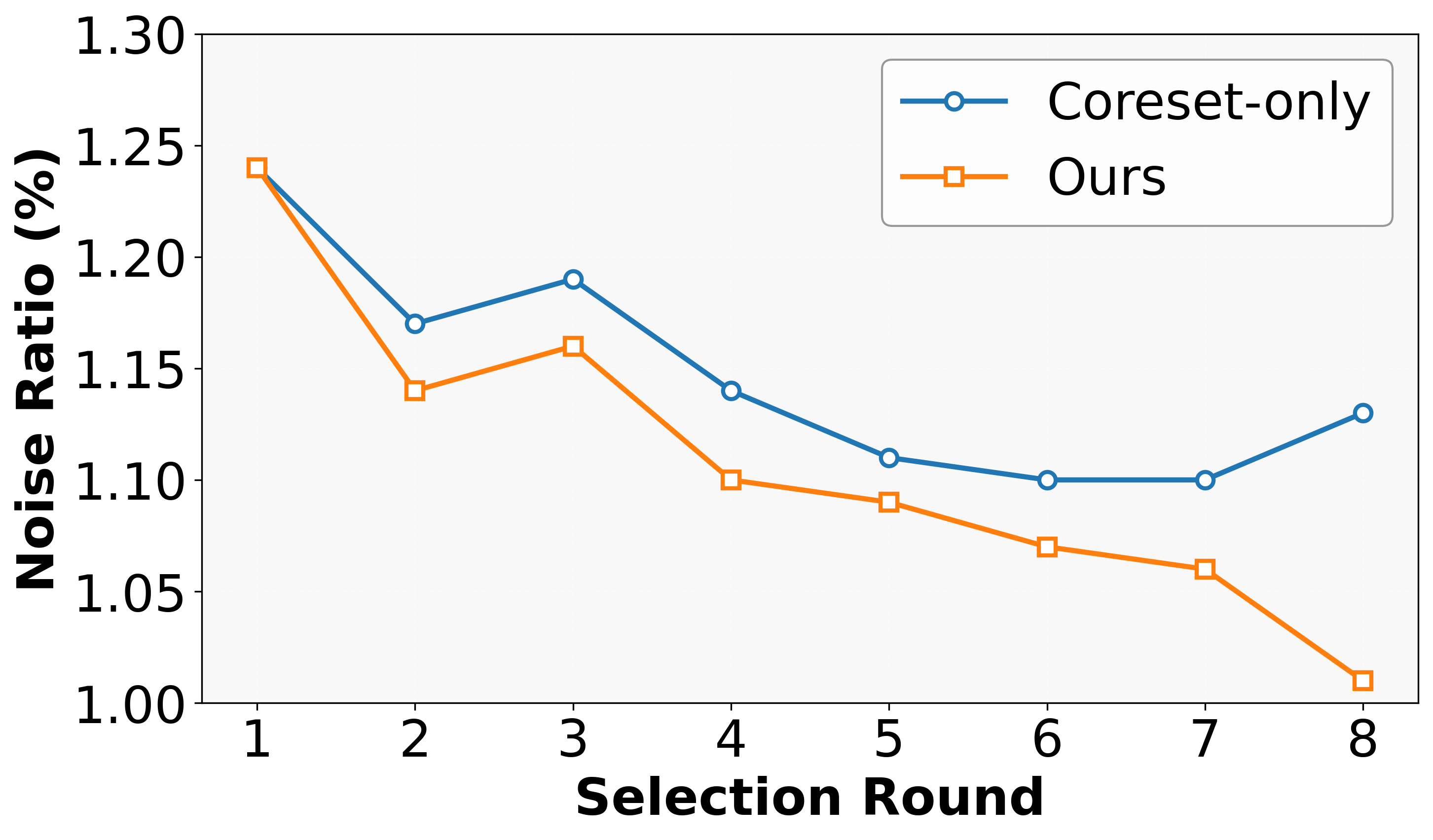}
    \caption{Noise in selected coresets}
    \label{fig:noise_ratio}
  \end{subfigure}
  \hfill
  \begin{subfigure}{0.32\linewidth}
    \includegraphics[width=1.0\linewidth]{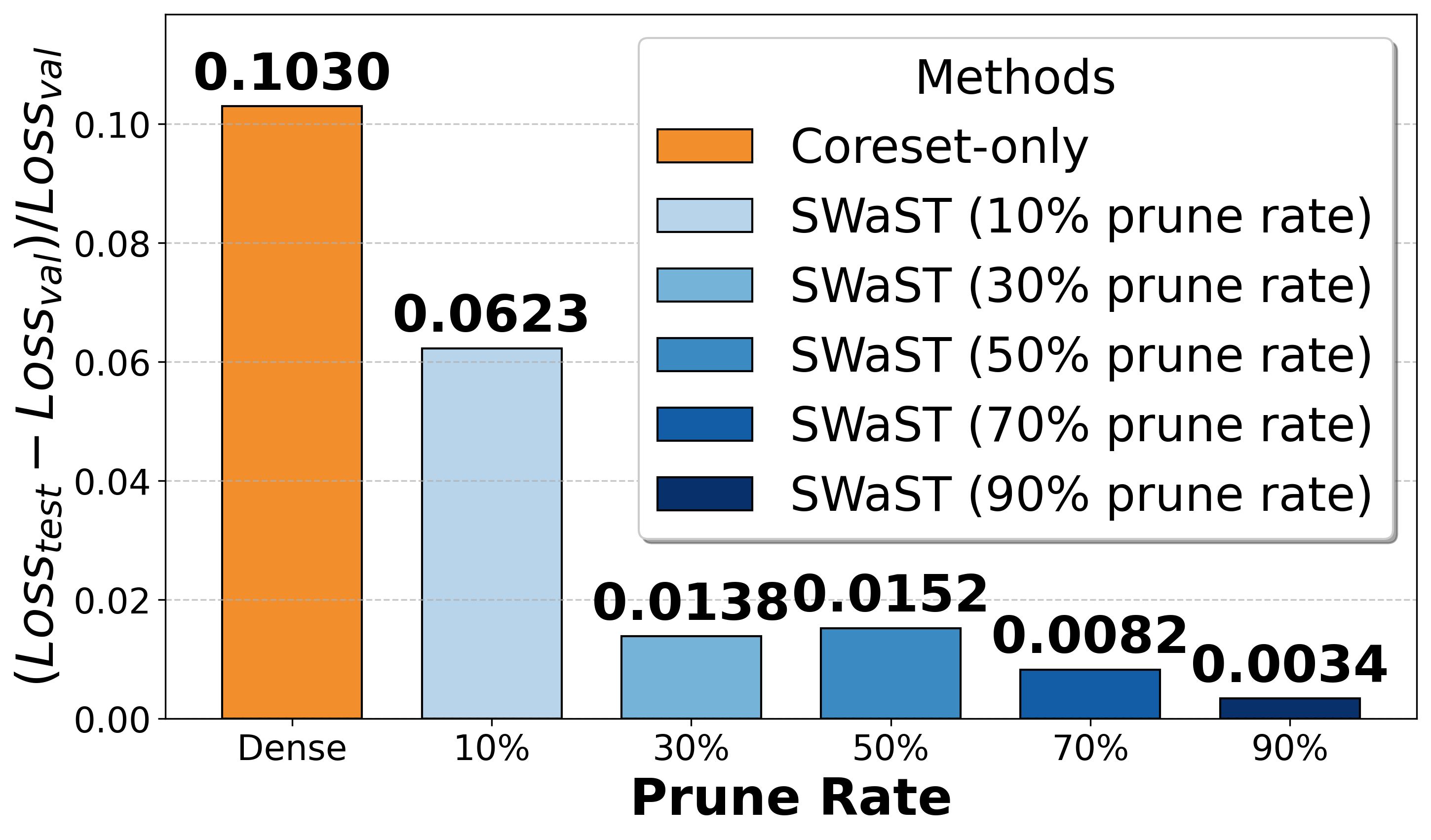}
    \caption{Overfitting}
    \label{fig:overfitting}
  \end{subfigure}
  \hfill
  \begin{subfigure}{0.32\linewidth}
    \includegraphics[width=1.0\linewidth]{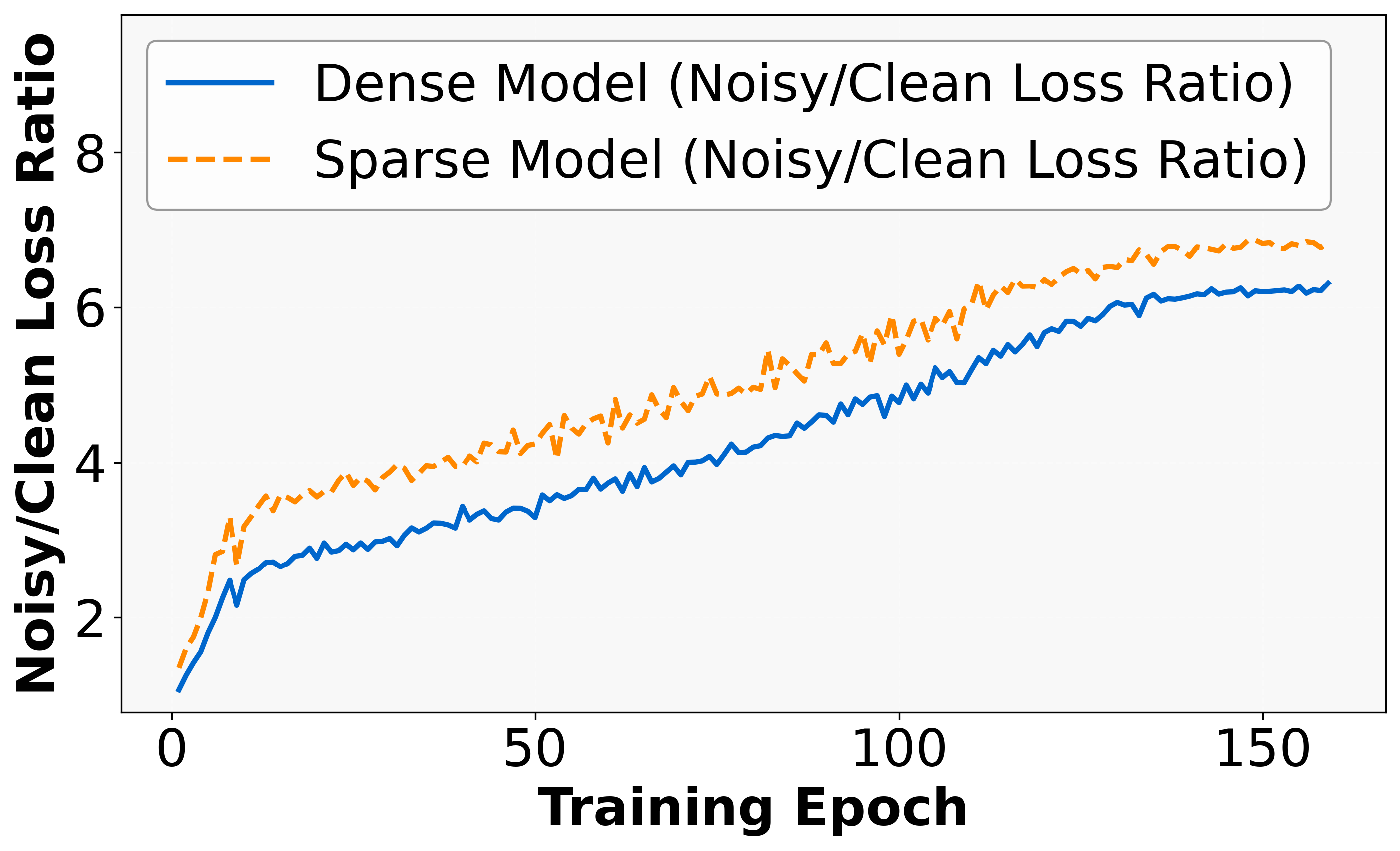}
    \caption{Loss differences}
    \label{fig:noise_loss}
  \end{subfigure}
  \caption{Results demonstrating the effectiveness of SWaST (lower values are better in (a) and (b), higher values are better in (c)).  
  (a) Noise ratio in selected coresets across training rounds comparing SWaST to the coreset-only baseline, with a 10.62\% reduction in the final selection. 
  (b) Overfitting comparison measured by test loss minus validation loss. SWaST significantly reduces overfitting compared to the dense model (Coreset-only), with higher prune rates yielding progressively better overfitting reduction.
  (c) Loss ratio between noisy and clean samples during training, showing that weight pruning increases the loss on noisy samples while preserving performance on clean data.}  
\end{figure*}
\begin{itemize}
    \item \textbf{Row-wise analysis:} Both SWaST-trim and SWaST-cut consistently outperform the coreset-only baseline across all configurations, achieving accuracy improvements of up to 17.83\%.
    \item \textbf{Column-wise analysis:} The performance gains from SWaST become more pronounced as coreset size diminishes. This trend aligns with our intuition: smaller coresets present greater selection challenges, where weight pruning facilitates the identification of higher-quality samples (see Fig.~\ref{fig:noise_ratio} and~\ref{fig:noise_loss}). Simultaneously, reduced coreset sizes exacerbate overfitting, which pruning effectively mitigates (see Fig.~\ref{fig:overfitting}).
\end{itemize}
Additional experiments on noisy datasets corroborate these findings, showing consistent performance improvements across diverse data conditions (see Appendix).

\begin{figure}[!htbp]
  \centering
  \begin{subfigure}{0.49\linewidth}
    \includegraphics[width=1.0\linewidth]{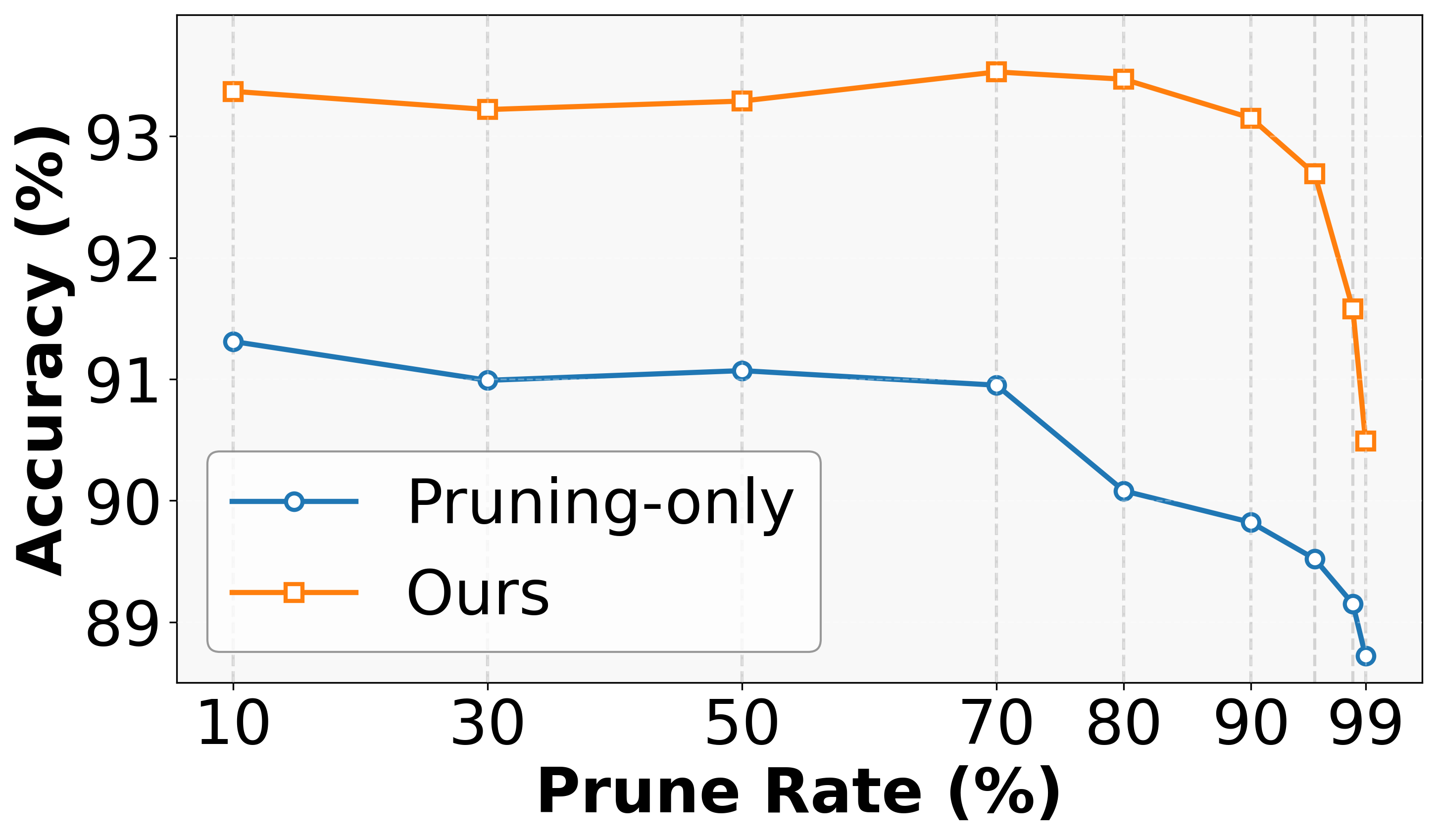}
    \caption{Accuracy vs. pruning rate}
    \label{fig:noise_prune}
  \end{subfigure}
  \hfill
  \begin{subfigure}{0.49\linewidth}
    \includegraphics[width=1.0\linewidth]{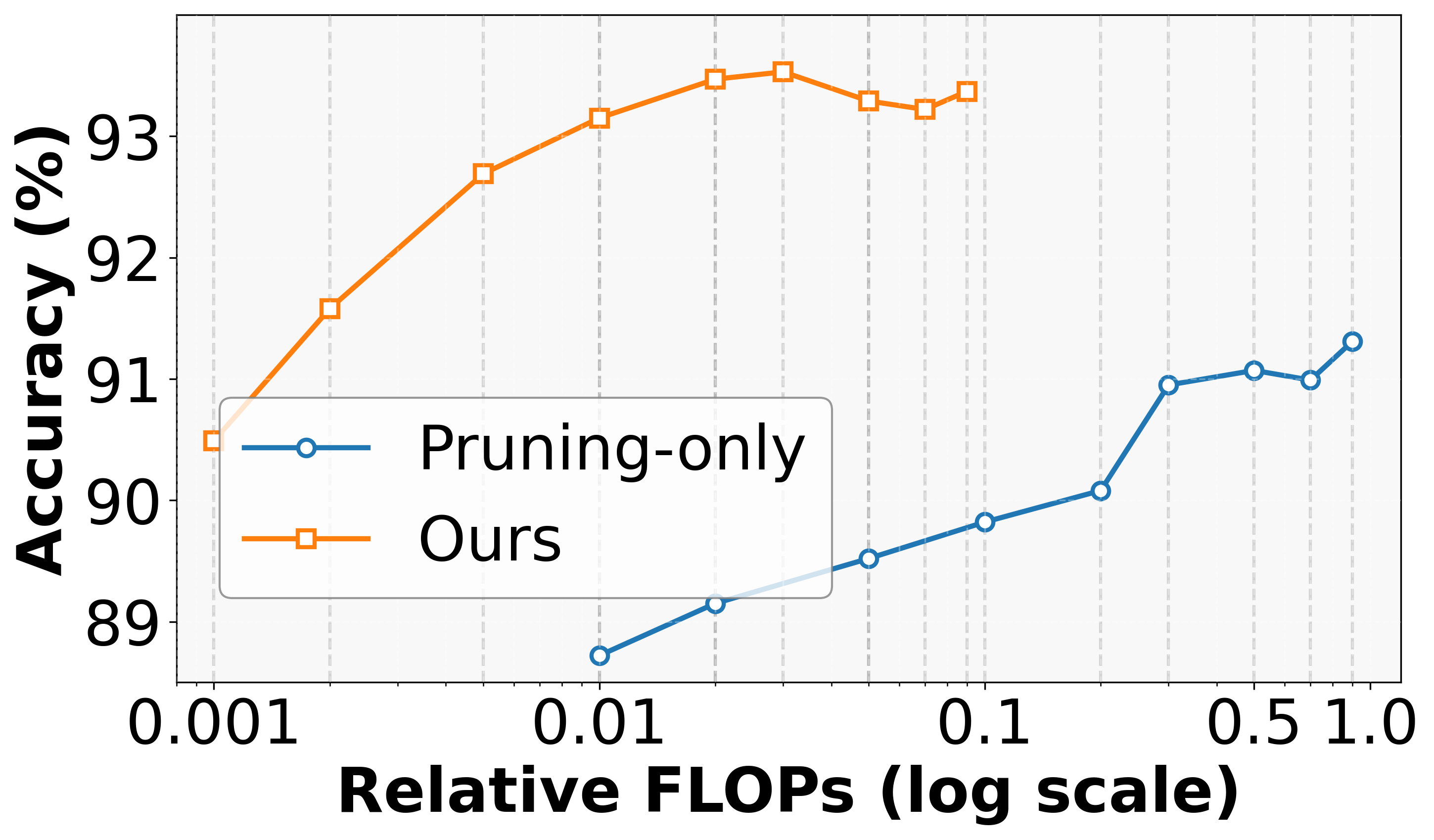}
    \caption{Accuracy vs. FLOPs}
    \label{fig:noise_prune_flops} 
  \end{subfigure}
  \caption{Performance comparison between SWaST-cut and pruning-only baseline. (a) SWaST-cut achieves higher accuracy across different pruning rates, with more pronounced improvements at high sparsity levels. (b) Due to coreset selection, SWaST-cut demonstrates superior accuracy-efficiency trade-offs, significantly outperforming the baseline under equivalent computational budgets.}
\end{figure}

\paragraph{Coreset selection improves weight pruning.}
We adopt the coreset algorithm from \citep{xia2022moderate} to identify the most informative samples. As shown in Fig.~\ref{fig:noise_prune} and Fig.~\ref{fig:noise_prune_flops}, SWaST delivers two key benefits:
\begin{itemize}
  \item \textit{Higher accuracy}: SWaST achieves 93.15\% accuracy at a 90\% pruning rate—an improvement of 3.33\% over the pruning-only baseline.
  \item \textit{Better efficiency}: Benefiting from the reduced training overhead through coreset selection, SWaST achieves superior accuracy under comparable FLOPs budgets with a gap of up to 4.43\% when the relative FLOPs is 0.01.
\end{itemize}

\section{Ablation Study}  
In the following, we conduct comprehensive ablation studies to analyze the key components of our framework. 

\paragraph{SWaST improves coreset quality.} Extending the experiments in Tab.~\ref{tab:tab1}, we further evaluate SWaST by introducing noise into the datasets. Using EL2N~\cite{paul2021deep} as the coreset selection method (detailed in the appendix), we track the proportion of noisy samples in each selected coreset throughout the training process. The results are visualized as line plots in Fig.~\ref{fig:noise_ratio}. The comparison shows that SWaST-cut helps the coreset algorithm better identify clean samples compared to the coreset-only training baseline. Specifically, in the final coreset, we achieves a 10.62\% reduction in the noise ratio relative to the baseline.

\paragraph{SWaST mitigates overfitting on coreset training.} 
Training on small coresets poses inherent overfitting risks due to limited data diversity. We evaluate this phenomenon by comparing test and validation loss between coreset-only training and our SWaST approach. As shown in Fig.~\ref{fig:overfitting}, SWaST consistently reduces the overfitting gap across all prune rates, with the reduction effect strengthening progressively as prune rate increases. This demonstrates that our sparse training paradigm effectively regularizes the model by preventing over-specialization to the limited coreset data. 

\paragraph{SWaST trains models with superior noise resistance.}
Following our previous setup, we extend the experiments from Tab.~\ref{tab:tab1} by introducing 10\% label noise into the datasets. We analyze the loss distributions of both noisy and clean samples throughout the training process, visualizing the results as line plots in Fig.~\ref{fig:noise_loss}. The visualization reveals a clear pattern: models with pruning exhibit higher loss values for noisy samples, indicating that pruning prevents the model from memorizing erroneous patterns in noisy data.

\section{Conclusion}
We explore the intricate interplay between weight pruning and coreset selection in deep neural networks, uncovering how redundant samples and weights can mutually undermine optimization efficiency and lead to challenges like the double-loss phenomenon. To overcome these issues, we introduce SWaST, an integrated training mechanism that enables simultaneous weight and sample tailoring through iterative compression and a state preservation mechanism.
Our extensive experiments reveal a strong synergy between pruning and coreset selection across varying prune rates and coreset sizes, which pave the way for more efficient deep learning paradigms.

\section*{Acknowledgments}
This work was supported by the National Nature Science Foundation of China (62472097), High-Quality Development Project of Shanghai Municipal Commission of Economy and Informatization (Grant No. 2024-GZL-RGZN-02010), and AI for Science Foundation of Fudan University (FudanX24AI028). The computations in this research were performed using the CFFF platform of Fudan University.

\bibliography{aaai2026}
\clearpage
\setcounter{page}{1}
\appendix

\section*{Appendix}
In this appendix, we provide comprehensive additional materials to supplement the main text. The contents include:
\begin{itemize}
\item \textbf{Broader impacts}: A discussion on the broader implications of our research, including enhanced training efficiency, environmental sustainability, and economic accessibility for smaller enterprises and research institutions.

\item \textbf{Experimental settings and details}: Comprehensive experimental setup including: \begin{itemize}
    \item Settings for interplay mechanism exploration
    \item Settings for primary experiments
    \item Settings for further analysis
    \item Settings for ablation studies
\end{itemize}

\item \textbf{Background: pruning methods}: Overview of the pruning methods used in our experimental evaluation.  

\item \textbf{Background: coreset selection methods}: Overview of the coreset selection methods utilized in our experimental evaluation.

\item \textbf{Analysis of backbone and FC layer pruning rates}: Investigation of performance under different pruning rates of the backbone and FC layer.  

\item \textbf{Detailed result on noisy data}: Investigation of SWaST's performance under noisy data conditions, validating its resilience and stability in challenging scenarios.

\item \textbf{Result on time efficiency}: Analysis of computational costs and training time improvements achieved through our method compared to traditional approaches. 

\item \textbf{Generalization}: Experimental validation of SWaST's compatibility with various pruning and coreset selection methods, demonstrating its effectiveness as a general-purpose approach.

\item \textbf{Limitations and future work}: Analysis of the limitations in our current work.

\end{itemize}

\section{Broader impacts}
\label{appendix:impacts}
The proposed integrated approach of alternating between network parameter pruning and subset selection not only enhances training efficiency but also promotes environmental sustainability by reducing energy consumption. Economically, it lowers the financial barriers for deploying advanced AI technologies, making them accessible to smaller enterprises and research institutions. By fostering more efficient and sustainable machine learning practices, our work contributes to the broader goal of democratizing AI and mitigating the environmental impact of large-scale computational tasks.

\section{Experimental settings and details}
\label{appendix:exp_details}
This section provides a comprehensive overview of the experimental configurations employed throughout our study. Unless otherwise specified, all experiments employ RigL~\cite{evci2020rigging} as the default pruning algorithm, following its original hyperparameter settings. We detail the settings for different experimental phases, including the exploration of the interaction mechanisms between pruning and coreset selection, primary performance evaluations, extended analyses, and ablation studies. For each set of experiments, we specify the datasets, model architectures, and hyperparameters to ensure reproducibility and clarity. All experiments were conducted on a server equipped with dual Intel Xeon Platinum 8369B CPUs (2.90GHz, 128 threads in total), 2.0 TiB RAM, and NVIDIA A100-SXM4 GPU with 80 GB memory (driver version 470.199.02, CUDA 12.6), running Ubuntu 22.04.4 LTS with Linux kernel 3.10.0-1160.el7.x86\_64.

\subsection{Settings for interplay mechanism exploration}
\paragraph{[Polynomial interpolation experiment]} To investigate the interaction between data selection and model pruning, we design the following polynomial fitting experiment.
\begin{itemize}
    \item \textbf{Number of Data Points (\(n\_points\))}: 10 data points were used in the experiment.
    \item \textbf{True Function}: The function used is \( y = \frac{1}{1 + 25x^2} \).
    \item \textbf{Noise Level}: Gaussian noise with a standard deviation of 0.1 was added to the data points.
    \item \textbf{Fraction of Noisy Data Points (\(r\))}: 50\% of the data points were subjected to noise.
    \item \textbf{Polynomial Degree (\(n\))}: A \(10^{\text{th}}\)-degree polynomial was fitted to the data.
    \item \textbf{Number of Coefficients to Retain (\(k\))}: The smallest 3 coefficients (in magnitude) were set to zero.
\end{itemize}

\paragraph{[Epsilon calculation experiment]}
To validate the \(\epsilon\) calculation in our theoretical analysis, we construct the experiment as follows.
\begin{itemize}
    \item \textbf{Data Generation Parameters:}
    \begin{itemize}
        \item \textbf{Number of Noisy Data Points (\(n1\))}: 50.
        \item \textbf{Number of Clean Data Points (\(n2\))}: 20.
        \item \textbf{Number of Test Data Points (\(m\))}: 100.
        \item \textbf{Noise Level}: Gaussian noise with a standard deviation of 0.1 was added to the noisy data points.
    \end{itemize}
    \item \textbf{Polynomial Fitting:}
    \begin{itemize}
        \item \textbf{Polynomial Degrees (\(k\))}: Ranges from 1 to 20.
    \end{itemize}
    \item \textbf{Loss Calculation:} Mean squared error (MSE) between the true function values and the polynomial predictions.
\end{itemize}

\subsection{Settings for primary experiments}
Our experiments are designed to demonstrate the bidirectional benefits between pruning and coreset selection:
\paragraph{[Settings for studying weight pruning's impact on coreset selection]} To comprehensively evaluate how weight pruning enhances coreset selection performance, we conduct experiments across multiple datasets using different coreset selection methods: GradMatch~\cite{killamsetty2021grad} for CIFAR-10/100, Moderate~\cite{xia2022moderate} for TinyImageNet, and EL2N~\cite{paul2021deep} for ImageNet-1K. We employ both ResNet-18 and ResNet-101 architectures to verify the effectiveness across different model scales. In terms of pruning configuration, we maintain a fixed 90\% pruning rate for FC layers while varying the backbone pruning rates according to specific experimental requirements. For training, we use SGD optimizer with momentum 0.9, initial learning rate 0.05, weight decay 5e-4, and nesterov momentum. The learning rate follows a cosine annealing schedule. We train for 300 epochs on CIFAR and TinyImageNet datasets, and 90 epochs on ImageNet-1K. The batch size is set to 128 for CIFAR datasets, 256 for TinyImageNet, and 512 for ImageNet-1K. Following GradMatch~\cite{killamsetty2021grad}, the warmup training period is set as $T_w = \left\lceil\frac{T\alpha}{2}\right\rceil$ epochs and selection interval $R=20$ epochs for CIFAR and TinyImageNet datasets, while for ImageNet-1K, we use $R=5$ epochs. These training configurations remain consistent throughout all subsequent experiments unless otherwise specified.

\paragraph{[Settings for studying coreset selection's impact on weight pruning]} To investigate how coreset selection benefits the pruning process, we follow the default RigL settings with pruning rates ranging from 10\% to 99\%, maintaining the same update frequency and sparsity distribution as described in~\cite{evci2020rigging}. We conduct experiments on CIFAR-10 using ResNet-18 architecture, with Moderate~\cite{xia2022moderate} as our coreset selection method and fixing the coreset size at 10\% of the original training set. To evaluate performance under challenging conditions, we additionally introduce random label corruption to 10\% of the training labels, allowing us to assess our framework's robustness comprehensively.

\subsection{Settings for further analysis}
To gain deeper insights into our framework's behavior, we design two sets of detailed experiments:
\paragraph{[Settings for studying coreset quality under noise]}
To investigate how our framework handles noisy data, we conduct experiments using CIFAR-10 with ResNet-18 architecture, employing EL2N~\cite{paul2021deep} as our coreset selection method. We intentionally corrupt 10\% of the training labels with random noise and set the coreset size to 10\% of the original training set. For the pruning configuration, we maintain a 90\% pruning rate throughout the network. We track the proportion of noisy samples in the selected coreset at each selection step to evaluate the framework's noise discrimination capability.

\paragraph{[Settings for analyzing noise sample discrimination]}
Following our primary experimental settings described earlier, we conduct this analysis on CIFAR-100 using ResNet-101 architecture, with an additional 10\% random label noise injection into the training set. We maintain all other training configurations from our primary experiments while monitoring the loss values for both clean and noisy samples throughout the training process. This setup allows us to examine how weight pruning affects the model's learning behavior with respect to different types of samples.

\subsection{Settings for ablation studies}
\paragraph{[Settings for analyzing different pruning configurations]}
To investigate the differential impacts of pruning different network components, we conduct experiments on CIFAR-10 using ResNet-18 architecture. We explore a comprehensive grid of pruning rates, varying both backbone pruning (from 10\% to 90\% with 20\% intervals) and FC layer pruning (from 10\% to 90\% with 20\% intervals). The coreset size is set to 1\% of the original training data. We maintain all other training configurations from our primary experiments, including optimizer settings and learning rate schedule.

\paragraph{[Settings for state preservation analysis]}  
For evaluating different state preservation strategies, we compare three variants: No Constraint (baseline), Selective Preservation (SP), and Complete Preservation (CP). These experiments are conducted using the same architecture (ResNet-101) and dataset (CIFAR-10 with 20\% noise), with Moderate for coreset selection and RigL for weight pruning. For joint optimization scenarios, we maintain a pruning ratio of 0.7 and coreset size of 10\%. To ensure statistical significance, all experiments are repeated 5 times with different random seeds, and we report both mean accuracy and standard deviation. Additional experiments applying state preservation to individual strategies (prune-only and coreset-only) use the same configuration to ensure fair comparison.  

\section{Background: pruning methods}
\label{appendix:used_pruning_method}
\subsection{RigL (Rigging the Lottery)}
RigL is a dynamic sparse training method that maintains network sparsity by periodically updating connections during training. The method consists of the following key components:

\paragraph{Parameter definition} Given a neural network $f_{\Theta}(\cdot)$ with parameters $\Theta \in \mathbb{R}^N$, where $\Theta^l$ (of length $N^l$) represents parameters of the $l$-th layer. Under sparse settings, each layer maintains a sparsity of $s^l \in (0,1)$, with actual parameter vector $\theta^l$ of length $(1-s^l)N^l$. The overall network sparsity is defined as $S=\frac{\sum_l s^lN^l}{N}$.

\paragraph{Sparsity distribution} RigL supports three strategies for distributing sparsity:
\begin{itemize}
    \item \textit{Uniform}: All layers except the first maintain the same sparsity $S$. The first layer remains dense due to its disproportional impact on performance.
    \item \textit{Erdős-Rényi}: Layer sparsity scales with $1- \frac{n^{l-1}+n^{l}}{n^{l-1} * n^{l}}$, where $n^l$ denotes the number of neurons in layer $l$.
    \item \textit{ERK (Erdős-Rényi-Kernel)}: For convolutional layers, sparsity scales with $1- \frac{n^{l-1}+n^{l}+w^l+h^l}{n^{l-1} * n^{l}*w^l*h^l}$, incorporating kernel dimensions $w^l$ and $h^l$. In our experiments, we adopt ERK as the default distribution strategy due to its superior performance in handling convolutional layers and better parameter efficiency.
\end{itemize}

\paragraph{Dynamic update mechanism} The core of RigL lies in its periodic update mechanism:
\begin{itemize}
    \item \textbf{Update Schedule}: Connections are updated every $\Delta T$ steps until reaching $T_{end}$.
    \item \textbf{Drop Criterion}: Removes weights with smallest absolute values, with the number determined by a cosine annealing function:
    $$f_{decay}(t;~\alpha,~T_{end})=\frac{\alpha}{2}\left(1+cos\left(\frac{t\pi}{T_{end}}\right)\right)$$
    \item \textbf{Grow Criterion}: Activates connections with highest gradient magnitudes, initializing new connections to zero.
\end{itemize}

\subsection{ProbMask (Probabilistic Masking)}
ProbMask introduces a probabilistic framework for network sparsification that operates under a global sparsity constraint. The method transforms the discrete pruning problem into a continuous optimization problem through probability space.

\paragraph{Problem formulation} Given a neural network with weights $\boldsymbol{w}\in \mathbb{R}^n$ and corresponding binary masks $\boldsymbol{m}\in \{0,1\}^n$, where $m_i=0$ indicates weight $w_i$ is pruned. The network sparsification is formulated as:
\begin{gather}
\min_{\boldsymbol{w}, \boldsymbol{m}} ~\mathcal{L}(\boldsymbol{w}, \boldsymbol{m}):=\frac{1}{N}\sum_{i=1}^{N} \ell\left(h\left(\mathbf{x}_{i} ; \boldsymbol{w\circ m}\right), \mathbf{y}_{i}\right) \nonumber\\
s.t.~ \boldsymbol{w}\in \mathbb{R}^n, \|\boldsymbol{m}\|_1 \leq K \mbox{ and }  \boldsymbol{m}\in \{0,1\}^n \nonumber
\end{gather}
where $K = kn$ represents the target model size with remaining ratio $k$.

\paragraph{Probabilistic relaxation} ProbMask relaxes the discrete optimization problem by treating each mask element as a Bernoulli random variable $m_i \sim \operatorname{Bern}(s_i)$, where $s_i \in [0,1]$ represents the probability of keeping the weight. This transforms the problem into:
\begin{gather}
\min_{\boldsymbol{w}, \boldsymbol{s}} \displaystyle ~\mathbb{E}_{ p(\boldsymbol{m}|\boldsymbol{s})} ~\mathcal{L}(\boldsymbol{w}, \boldsymbol{m}) \nonumber\\ 
s.t. ~\boldsymbol{w}\in \mathbb{R}^n, \boldsymbol{1}^\top \boldsymbol{s}  \leq K \mbox{ and } \boldsymbol{s}\in [0,1]^n \nonumber
\end{gather}

\paragraph{Optimization strategy} The method employs projected gradient descent with the following key components:
\begin{itemize}
    \item \textbf{Gradient Computation}: Utilizes the Gumbel-Softmax trick to calculate gradients with respect to both weights and probabilities.
    \item \textbf{Temperature Annealing}: Applies a decreasing temperature parameter $\tau$ during training to encourage probability convergence.
    \item \textbf{Gradual Sparsification}: Increases pruning rate smoothly using the cubic function:
    $$k=k_{f}+\left(1-k_{f}\right)\left(1-\frac{t-t_{1}}{t_{2}-t_{1}}\right)^{3}$$
\end{itemize}

\section{Background: coreset selection methods}
\label{appendix:used_selection_method}
\subsection{GradMatch}
GradMatch is an adaptive data selection algorithm that aims to select a subset of training data by minimizing the gradient matching error between the selected subset and the full dataset.

\paragraph{Core Objective} The method optimizes the subset selection by minimizing the error between the weighted gradients of the selected subset and the full dataset gradients, with an additional L2 regularization term to prevent overfitting.

\paragraph{Optimization strategy}
The optimization is solved using Orthogonal Matching Pursuit (OMP), which:
\begin{itemize}
    \item Iteratively selects data points that maximize gradient matching
    \item Provides a weakly submodular optimization with approximation guarantees
    \item Terminates when either reaching the target subset size or achieving error tolerance
\end{itemize}

\subsection{Moderate}
Moderate is a universal data selection method that aims to select representative samples by considering the moderate distance between data points and their class centers. Unlike previous methods that target specific scenarios, Moderate is designed to be robust across various real-world situations.

\paragraph{Core principle} The method is based on the concept of "moderate coreset", which selects data points with scores close to the median score rather than extreme values. For any score criterion, different scenarios prefer data points with scores in different intervals. The score median serves as a statistical proxy of the full data distribution.

\subsection{EL2N}
EL2N (Expected L2-Norm) is a data selection method that scores training samples based on their prediction error magnitude during early training stages.

\paragraph{Core principle} 
The EL2N score for a training sample $(x,y)$ is defined as $E\|p(w_t,x)-y\|^2$, where $p(w_t,x)$ is the network's prediction probability vector at time $t$, and $y$ is the one-hot label vector.

\paragraph{Modified strategy}
In our implementation, we propose an adaptive selection strategy based on training progress. While EL2N is particularly effective during early training stages, we modify the selection criterion at the midpoint of training. Specifically, after reaching 50\% of the total training epochs, we switch to preferentially selecting samples with lower EL2N scores. This modification acknowledges that as the model becomes more mature in its training process, samples with lower prediction errors are more valuable for fine-tuning the model's performance.

\subsection{GLISTER}
GLISTER is a data selection framework that aims to select a subset of training data by maximizing the log-likelihood on a held-out validation set. It formulates the selection as a mixed discrete-continuous bi-level optimization problem.

\paragraph{Core objective} 
The method optimizes for:
\begin{equation}
S^* = \arg\max_{S\subseteq U,|S|\leq k} LL_V(\arg\max_{\theta} LL_T(\theta,S), V)
\end{equation}
where $LL_V$ and $LL_T$ are log-likelihood functions on validation set $V$ and training set $U$ respectively, and $k$ is the target subset size.

\subsection{CRAIG}
CRAIG is a data selection framework that formulates subset selection as a facility location problem, aiming to approximate the full gradient with a weighted subset of training points.

\paragraph{Core objective} 
The method solves the optimization problem:
\begin{equation}
\min_{S\subseteq V, |S|\leq k} \max_{w\geq 0} \|\sum_{i\in V}\nabla f_i(w_t) - \sum_{j\in S}w_j\nabla f_j(w_t)\|
\end{equation}
where $f_i$ represents the loss for sample $i$, $w_t$ is the model parameter at time $t$, and $k$ is the target subset size.

\section{Analysis of backbone and FC layer pruning}
\label{sec:prune_ablation}
\begin{figure}[!htbp]  
  \centering  
  \includegraphics[width=0.8\linewidth]{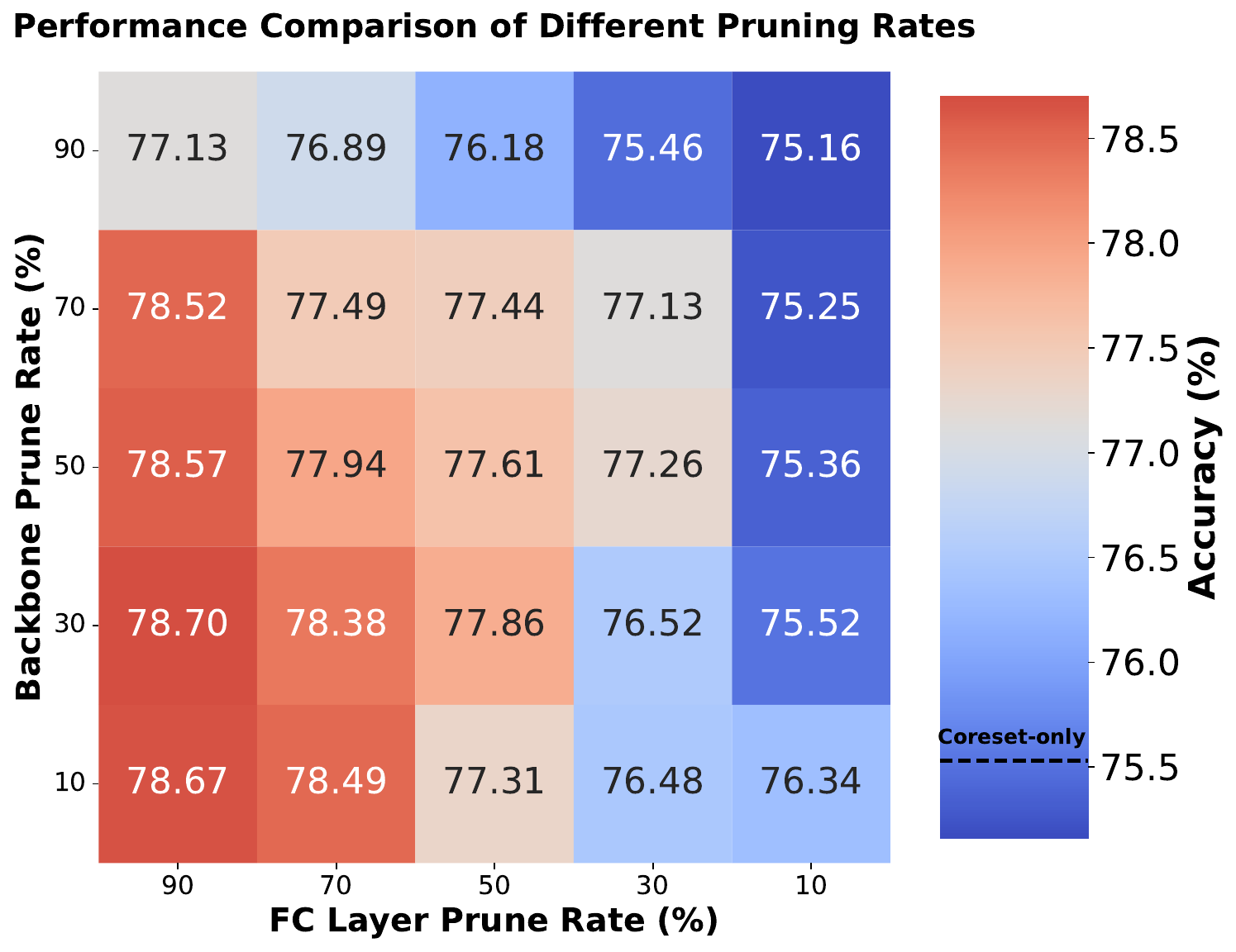}  
   \caption{Performance analysis across different pruning configurations. The x-axis and y-axis represent the pruning rates for FC layers and backbone respectively. Each cell shows the accuracy (\%). The horizontal dashed line indicates the baseline accuracy (75.53\%) from coreset-only. Higher FC layer pruning rates (\(>\)70\%) consistently yield better performance.}  
   \label{fig:prune-ablation}  
\end{figure}

Experiments reveal asymmetric contributions of network components to pruning-coreset synergy. While our framework outperforms coreset-only training with reduced total FLOPs, the performance gains primarily stem from FC layer pruning, with backbone pruning playing a secondary role (as shown in Fig.~\ref{fig:prune-ablation}).
To investigate this phenomenon, we conduct controlled experiments with varying pruning configurations. Throughout our main experiments, we maintain a fixed 90\% pruning rate for FC layers unless otherwise specified, with the term “pruning rate" referring specifically to backbone pruning. This configuration choice is based on our ablation results, which demonstrate that FC layer pruning provides the most significant performance benefits.

\section{Discussion on state preservation}
\label{appendix:state_preservation}
Our state preservation mechanism fundamentally differs from traditional knowledge distillation (KD). KD transfers knowledge from a teacher to a student model, while our approach preserves the model's historical states during alternating optimization, retaining key information from pruned parameters and reduced datasets to mitigate degradation.

Unlike KD's external teacher-student setup, our mechanism is an internal constraint for co-optimizing weight pruning and coreset selection. Thus, comparisons with KD-based methods that use external models for distillation to enhance post-pruning model performance are inappropriate due to differing objectives and assumptions. By avoiding external models or soft targets, we address unique coreset/pruning challenges without biases, ensuring fair evaluation.
Ablation studies in the main text validate this: applying state preservation to coreset-only or pruning-only scenarios shows no gains, demonstrating it stabilizes joint optimization, not enhances performance like KD.

\section{More result on time efficiency}
\label{appendix:time_efficiency}
To thoroughly evaluate the practical benefits of our framework, we conduct a detailed analysis of its time efficiency in comparison with baseline methods. As shown in Table~\ref{tab:time}, our method with 10\% coreset size achieves comparable accuracy (92.96\%) to the coreset-only method with 15\% data (92.97\%), while requiring significantly less training time (2191s vs 2738s).
\begin{table}[!htbp]  
\footnotesize   
\centering  
\caption{\textbf{Comparison of accuracy and time efficiency.} }   
\label{tab:time}  
\resizebox{\linewidth}{!}{   
\begin{tabular}{@{}cccc@{}}  
\toprule  
\textbf{Settings}  & coreset-only (10\%) & coreset-only (15\%) & \textbf{Ours (10\%)} \\ \midrule  
\textbf{Acc. (\%) ↑}     & 91.13              & 92.97              & \textbf{92.96}      \\
\textbf{Time (sec) ↓} & 2034              & 2738              & \textbf{2191}      \\
\textbf{Params (M) ↓} & 44.5              & 44.5              & \textbf{31.2}      \\ \bottomrule  
\end{tabular}}   
\end{table} 
Moreover, when compared to pruning-only methods using full dataset, ours provides significant speed advantage by using coreset.

\begin{table*}[!htbp] 
    \caption{Comparison of test accuracy (\%) on CIFAR-10 across different training settings. The table is divided into four distinct regions: (1) Standard training ({\color{gray!85}gray}, top-right cell) achieves 85.99\% accuracy with full dataset; (2) Pruning-only training ({\color{orange!85}orange}, top row) shows accuracy under different pruning rates, ranging from -0.14\% to +0.87\%; (3) Coreset-only training ({\color{blue!85}blue}, rightmost column); and (4) Our joint pruning and coreset selection framework ({\color{green!85}green}) demonstrates superior performance with up to +6.12\% improvement over standard training. Values in parentheses indicate differences from the Dense training baseline.} 
    \label{tab:noise_result}
    \centering  
    \resizebox{\textwidth}{!}{  
    \renewcommand{\arraystretch}{1.2}
    \begin{tabular}{lcccccc}  
    \hline  
    & \multicolumn{6}{c}{\textbf{Prune Rate in Joint Weight and Sample Tailoring}} \\ \cline{2-7}  
    \multirow{-2}{*}{\textbf{Subset Fraction}} & 90\% & 70\% & 50\% & 30\% & 10\% & 0\% (Dense) \\ \hline  
    100\% ({\color{orange!85}Pruning-only}) & \cellcolor{orange!15}$85.85_{{\tiny(-0.14)}}$ & \cellcolor{orange!15}$86.12_{{\tiny(+0.13)}}$ & \cellcolor{orange!15}{$86.41_{{\tiny(+0.42)}}$} & \cellcolor{orange!15}{$86.86_{{\tiny(+0.87)}}$} & \cellcolor{orange!15}$86.70_{{\tiny(+0.71)}}$ & \cellcolor{gray!15}$85.99$ (Standard Training) \\ 
    ~20\%~~({\color{green!85}Ours}) & \cellcolor{green!15}$91.08_{{\tiny(+0.75)}}$ & \cellcolor{green!15}$91.86_{{\tiny(+1.53)}}$ & \cellcolor{green!15}{$91.89_{{\tiny(+1.56)}}$} & \cellcolor{green!15}{$92.11_{{\tiny(+1.78)}}$} & \cellcolor{green!15}{$92.04_{{\tiny(+1.71)}}$} & \cellcolor{blue!15}$90.33$ (Coreset-only) \\
    ~10\%~~({\color{green!85}Ours}) & \cellcolor{green!15}$87.82_{{\tiny(+1.01)}}$ & \cellcolor{green!15}$88.37_{{\tiny(+1.56)}}$ & \cellcolor{green!15}{$88.62_{{\tiny(+1.81)}}$} & \cellcolor{green!15}{$88.69_{{\tiny(+1.88)}}$} & \cellcolor{green!15}{$88.64_{{\tiny(+1.83)}}$} & \cellcolor{blue!15}$86.81$ (Coreset-only) \\ 
    ~~5\%~~~({\color{green!85}Ours}) & \cellcolor{green!15}$85.14_{{\tiny(+8.47)}}$ & \cellcolor{green!15}$86.05_{{\tiny(+9.38)}}$ & \cellcolor{green!15}{$86.19_{{\tiny(+9.52)}}$} & \cellcolor{green!15}{$86.45_{{\tiny(+9.78)}}$} & \cellcolor{green!15}{$86.34_{{\tiny(+9.67)}}$} & \cellcolor{blue!15}$76.67$ (Coreset-only) \\ \hline 
    \end{tabular}}  
\end{table*}

\section{Detail results on noisy data}  
\label{appendix:noise_exp}  
To evaluate the robustness of our framework in practical scenarios, we conduct experiments with 20\% noise in training data. As shown in Table~\ref{tab:noise_result}, our method demonstrates remarkable performance even under such noisy conditions. With only 20\% of the training data, our joint framework achieves up to 92.11\% accuracy (+6.12\% improvement over standard training), significantly outperforming both pruning-only (maximum 87.46\%) and coreset-only (90.33\%) approaches. Notably, this superior performance is achieved with substantially reduced computational costs, using both fewer parameters (through pruning) and less training data (through coreset selection). Even when further reducing the data to just 5\%, our method maintains competitive performance (86.45\%), demonstrating its efficiency in handling noisy scenarios with minimal resources. These results highlight that our framework not only enhances model robustness against label noise but also achieves this with significantly lower computational overhead compared to traditional training approaches.

\begin{table*}[!htbp]   
    \caption{Test accuracy (\%) of our framework using ProbMask as the pruning method. The coreset size is fixed at 10\% of the training set. Values in parentheses indicate differences from the Coreset-only baseline. Red/blue numbers indicate best and runner-up results within each row.}   
    \label{tab:probmask}  
    \centering  
    \resizebox{\linewidth}{!}{  
    \renewcommand{\arraystretch}{1.1}  
    \begin{tabular}{ccccccccc}  
    \hline
    & &  & \multicolumn{5}{c}{\textbf{Prune Rate in Joint Weight and Sample Tailoring}} &  \\ \cline{4-8}  
    \multirow{-2}{*}{\textbf{Pruning Method}} & \multirow{-2}{*}{\textbf{Model}} & \multirow{-2}{*}{\textbf{Dataset}} & 90\% & 70\% & 50\% & 30\% & 10\% & \multirow{-2}{*}{\textbf{Coreset-only}} \\ \hline  
    & & CIFAR-10 & $91.47_{{\tiny(-0.73)}}$ & $92.31_{{\tiny(+0.11)}}$ & {\color{blue}$92.44_{{\tiny(+0.24)}}$} & {\color{red}\textbf{$92.71_{{\tiny(+0.51)}}$}} & {\color{blue}$92.59_{{\tiny(+0.39)}}$} & 92.20 \\   
    & \multirow{-2}{*}{ResNet-18} & CIFAR-100 & $67.74_{{\tiny(-3.23)}}$ & $71.03_{{\tiny(+0.06)}}$ & {\color{blue}$71.56_{{\tiny(+0.59)}}$} & {\color{red}\textbf{$71.93_{{\tiny(+0.96)}}$}} & {\color{blue}$71.87_{{\tiny(+0.90)}}$} & 70.97 \\ \cline{2-9}  
    & & CIFAR-10 & $91.93_{{\tiny(+0.80)}}$ & $92.28_{{\tiny(+1.15)}}$ & {\color{blue}$92.46_{{\tiny(+1.33)}}$} & {\color{red}\textbf{$92.88_{{\tiny(+1.75)}}$}} & {\color{blue}$92.76_{{\tiny(+1.63)}}$} & 91.13 \\   
    \multirow{-4}{*}{ProbMask} & \multirow{-2}{*}{ResNet-101} & CIFAR-100 & $69.18_{{\tiny(+0.39)}}$ & $71.05_{{\tiny(+2.26)}}$ & {\color{blue}$71.74_{{\tiny(+2.95)}}$} & {\color{red}\textbf{$72.19_{{\tiny(+3.40)}}$}} & {\color{blue}$71.96_{{\tiny(+3.17)}}$} & 68.79 \\ \hline   
    \end{tabular}}  
\end{table*}

\begin{table*}[!t]   
    \caption{Test accuracy (\%) of our framework using ProbMask as the pruning method. Each entry shows the test accuracy with the relative improvement over coreset-only baseline in subscript. The coreset size is fixed at 10\% of the training set. Red/blue numbers indicate best and runner-up results within each row.}   
    \label{tab:other_coreset_method}  
    \centering  
    \resizebox{\linewidth}{!}{  
    \renewcommand{\arraystretch}{1.1}  
    \begin{tabular}{cccccccc}  
    \hline
    &  & \multicolumn{5}{c}{\textbf{Prune Rate in Joint Weight and Sample Tailoring}} &  \\ \cline{3-7}  
    \multirow{-2}{*}{\textbf{Coreset Methods}} & \multirow{-2}{*}{\textbf{Model}} & 90\% & 70\% & 50\% & 30\% & 10\% & \multirow{-2}{*}{\textbf{Coreset-only}} \\ \midrule  
    & ResNet-18 & $90.87_{{\tiny(+1.54)}}$ & $91.41_{{\tiny(+2.08)}}$ & {\color{blue}$91.63_{{\tiny(+2.30)}}$} & {\color{red}\textbf{$91.77_{{\tiny(+2.44)}}$}} & {\color{blue}$91.75_{{\tiny(+2.42)}}$} & 89.33 \\   
    \multirow{-2}{*}{CRAIG} & ResNet-101 & $91.03_{{\tiny(+5.76)}}$ & $91.66_{{\tiny(+6.39)}}$ & {\color{blue}$91.80_{{\tiny(+6.53)}}$} & {\color{red}\textbf{$91.82_{{\tiny(+6.55)}}$}} & {\color{blue}$91.71_{{\tiny(+6.44)}}$} & 85.27 \\ \hline  
    & ResNet-18 & $91.42_{{\tiny(-0.83)}}$ & $92.50_{{\tiny(+0.25)}}$ & {\color{blue}$92.62_{{\tiny(+0.37)}}$} & {\color{red}\textbf{$92.85_{{\tiny(+0.60)}}$}} & {\color{blue}$92.79_{{\tiny(+0.54)}}$} & 92.25 \\
    \multirow{-2}{*}{Glister} & ResNet-101 & $91.82_{{\tiny(+0.36)}}$ & $92.61_{{\tiny(+1.15)}}$ & {\color{blue}$92.95_{{\tiny(+1.49)}}$} & {\color{red}\textbf{$93.12_{{\tiny(+1.66)}}$}} & {\color{blue}$93.03_{{\tiny(+1.57)}}$} & 91.46 \\ \hline  
    & ResNet-18 & $91.12_{{\tiny(+0.33)}}$ & $91.35_{{\tiny(+0.56)}}$ & {\color{blue}$91.58_{{\tiny(+0.79)}}$} & {\color{red}\textbf{$91.67_{{\tiny(+0.88)}}$}} & {\color{blue}$91.55_{{\tiny(+0.76)}}$} & 90.79 \\   
    \multirow{-2}{*}{Moderate} & ResNet-101 & $91.24_{{\tiny(+3.07)}}$ & $91.39_{{\tiny(+3.22)}}$ & {\color{blue}$91.65_{{\tiny(+3.48)}}$} & {\color{red}\textbf{$91.88_{{\tiny(+3.71)}}$}} & {\color{blue}$91.74_{{\tiny(+3.57)}}$} & 88.17 \\ \hline  
    \end{tabular}}  
\end{table*}

\section{Generalization}
\label{appendix:using_other_method}
As emphasized in our main text, a key advantage of our proposed framework lies in its flexibility and generalizability. Our framework is designed to accommodate a wide range of alternative approaches. This section demonstrates this generalizability through comprehensive experiments with different pruning algorithms and coreset selection methods. The following subsections present experimental results that validate our framework's compatibility with various methods, while maintaining consistent performance improvements.

\subsection{Generalization on pruning method}
To demonstrate the generality of our framework, we evaluate its performance using ProbMask~\cite{zhou2022probabilistic} as an alternative pruning method. As shown in Table~\ref{tab:probmask}, our framework maintains its effectiveness across different network architectures (ResNet-18 and ResNet-101) and datasets (CIFAR-10 and CIFAR-100). With ResNet-18, we achieve improvements of up to +0.51\% and +0.96\% over the coreset-only baseline on CIFAR-10 and CIFAR-100, respectively. The benefits become more pronounced with ResNet-101, where we observe larger gains of up to +1.75\% on CIFAR-10 and +3.40\% on CIFAR-100. Notably, the optimal performance consistently occurs at moderate pruning rates (around 30\%), regardless of the model architecture or dataset. These results suggest that our framework's ability to jointly optimize network structure and training data is not tied to specific pruning methods, demonstrating its potential as a general approach for efficient deep learning.

\subsection{Generalization on coreset selection method}
To further validate the versatility of SWaST, we evaluate its performance with different coreset selection methods, including CRAIG~\cite{mirzasoleiman2020coresets}, Glister~\cite{killamsetty2021glister}, and Moderate~\cite{xia2022moderate}. As shown in Table~\ref{tab:other_coreset_method}, our framework consistently improves upon the coreset-only baselines across all methods and architectures. With CRAIG, we achieve substantial improvements of up to +2.44\% and +6.55\% on ResNet-18 and ResNet-101, respectively. Similar patterns are observed with GLISTER (up to +0.60\% and +1.66\%) and Moderate (up to +0.88\% and +3.71\%). Notably, across all coreset methods, the optimal performance is consistently achieved at 30\% pruning rate, suggesting a robust sweet spot for the pruning-coreset trade-off. These results demonstrate that our framework's effectiveness is not limited to specific coreset selection algorithms, but rather provides a general approach for combining network pruning with data selection, regardless of the underlying coreset method.

\section{Limitations}
\label{appendix:limitation}
While our framework demonstrates promising results in joint optimization of network structure and training data, there are certain limitations in practical acceleration that warrant discussion. Although our approach theoretically enables multiplicative speedup - combining the acceleration from weight pruning ($a\times$) and coreset selection ($b\times$) - achieving the full $a \times b$ acceleration in practice remains challenging. This gap between theoretical and practical speedup primarily stems from the current limitations in implementing efficient sparse computation.
The main challenge lies in translating theoretical sparsity into practical acceleration through weight pruning. Modern deep learning frameworks and hardware are often optimized for dense computation, making it difficult to fully capitalize on the potential benefits of sparse networks without substantial engineering efforts. These efforts would include developing specialized sparse computation kernels, implementing efficient model instantiation mechanisms, and optimizing hardware-specific operations.

\end{document}